\documentclass[]{article}

\usepackage{arxiv}

\usepackage[utf8]{inputenc} 
\usepackage[T1]{fontenc}    
\usepackage{hyperref}       
\usepackage{url}            
\usepackage{booktabs}       
\usepackage{amsfonts}       
\usepackage{nicefrac}       
\usepackage{microtype}      
\usepackage{lipsum}
\usepackage{fancyhdr}       
\usepackage{graphicx}       
\graphicspath{{media/}}     
\usepackage{enumitem}
\usepackage{multirow}
\usepackage{amsmath}
\usepackage{setspace}
\usepackage[margin=10pt,font=small,labelfont=bf]{caption}
\usepackage{subcaption}

\newcommand{\centered}[1]{\begin{tabular}{l} #1 \end{tabular}}

\def \A {\mathcal A}
\def \T {\mathcal T}
\def \B {\mathcal B}
\def \D {\mathcal D}
\def \N {\mathcal N}

\usepackage[capitalize]{cleveref}
\crefname{section}{Sec.}{Secs.}
\crefname{section}{Section}{Sections}
\crefname{table}{Table}{Tables}
\crefname{table}{Tab.}{Tabs.}

\title{Back-to-Bones: Rediscovering the Role of Backbones in Domain Generalization}

\author{
  Simone Angarano$^{1}$, Mauro Martini$^{1}$, Francesco Salvetti$^{1,2}$, Vittorio Mazzia$^{1,2}$, Marcello Chiaberge$^{1}$\\
  $^1$PIC4SeR PoliTo Interdepartmental Center for Service Robotics\\
  $^2$SmartData@PoliTo, Big Data and Data Science Laboratory\\
  Department of Electronics and Telecommunications, Politecnico di Torino, Turin, Italy\\
  \texttt{\{name.surname\}@polito.it}
}

\begin{document}
\maketitle

\begin{abstract}
Domain Generalization (DG) studies the capability of a deep learning model to generalize to out-of-training distributions. In the last decade, literature has been massively filled with training methodologies that claim to obtain more abstract and robust data representations to tackle domain shifts. Recent research has provided a reproducible benchmark for DG, pointing out the effectiveness of naive empirical risk minimization (ERM) over existing algorithms. Nevertheless, researchers persist in using the same outdated feature extractors, and no attention has been given to the effects of different backbones yet. In this paper, we start back to the backbones proposing a comprehensive analysis of their intrinsic generalization capabilities, which so far have been ignored by the research community. We evaluate a wide variety of feature extractors, from standard residual solutions to transformer-based architectures, finding an evident linear correlation between large-scale single-domain classification accuracy and DG capability. Our extensive experimentation shows that by adopting competitive backbones in conjunction with effective data augmentation, plain ERM outperforms recent DG solutions and achieves state-of-the-art accuracy.
Moreover, our additional qualitative studies reveal that novel backbones give more similar representations to same-class samples, separating different domains in the feature space. This boost in generalization capabilities leaves marginal room for DG algorithms. It suggests a new paradigm for investigating the problem, placing backbones in the spotlight and encouraging the development of consistent algorithms on top of them. The code is available at \url{https://github.com/PIC4SeR/Back-to-Bones}.
\end{abstract}


\begin{figure}[t]
  \centering
   \includegraphics[width=250pt,
                    trim={5px 10px 5px 0px}]
                   {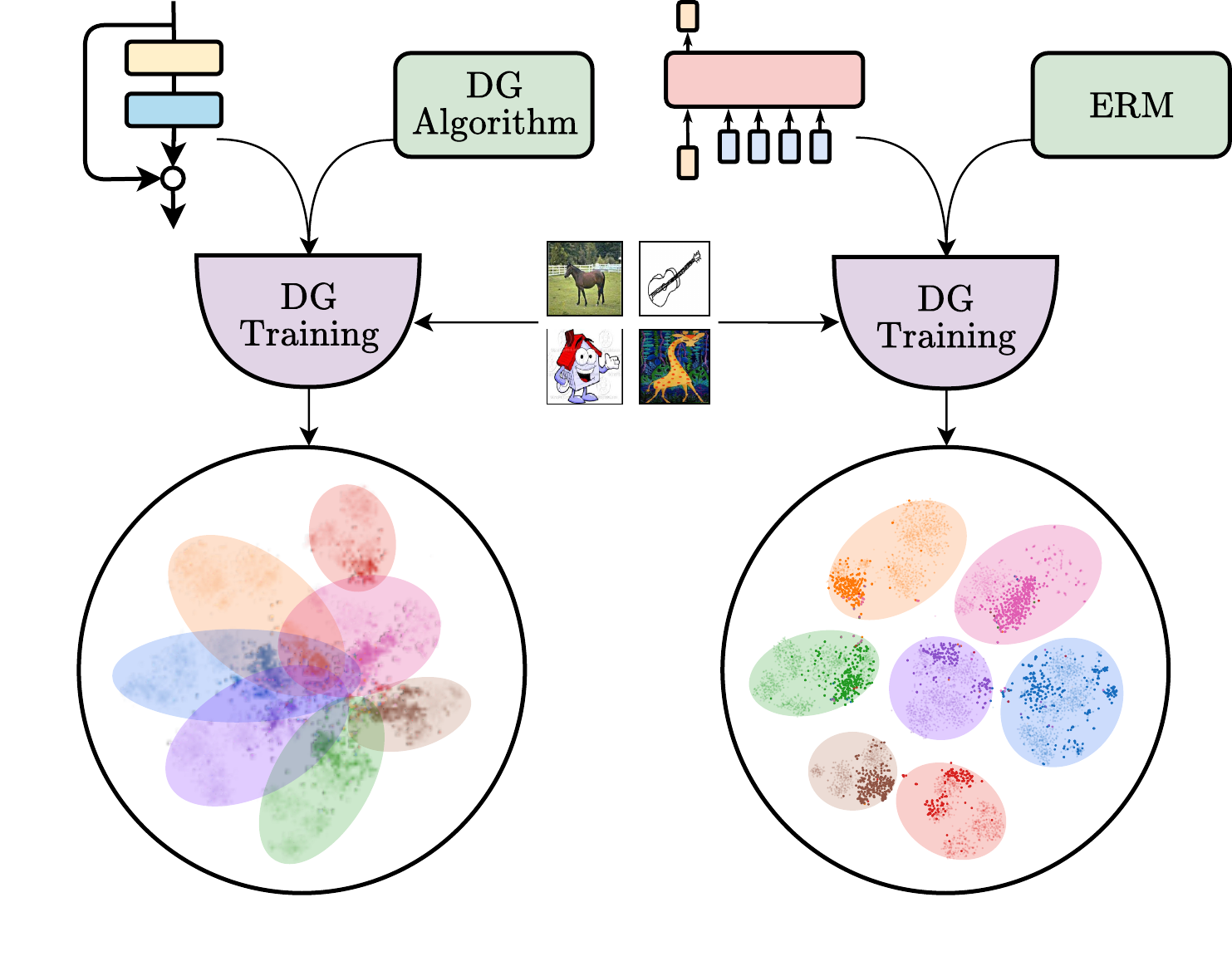}
   \caption{Our experimentation proves the importance of backbones in Domain Generalization. We find that novel architectures, such as transformed-based models, lead to a better representation of data, outperforming outdated backbones, such as ResNets, and leaving marginal room for feature mapping improvement using DG algorithms.}
   \label{fig:firstpage}
\end{figure}

\section{Introduction}
\label{sec:intro}
The problem of induction has a central role in the learning process. Without generalization, machine learning algorithms would be able to exhibit useful behaviors only in situations identical to the ones previously experienced \cite{valiant2013probably}. 
Deep neural networks are powerful models capable of extracting subtle regularities from training data. Nevertheless, they often fail to generalize to out-of-training data. Even if supervised training methodologies have proved to produce neural networks with remarkable performances, their results are valid only in well-defined settings and do not generalize across tasks, domains, and categories \cite{csurka2017domain}. For the specific object recognition task, several literature works have shown that, unlike humans, training frameworks commonly produce networks that are more prone to be biased towards textures and global image statistics in making decisions \cite{geirhos2018imagenet,gatys2015texture}, prioritizing easier-to-fit spurious correlations in favor of invariant shape cues \cite{arjovsky2019invariant}. That prevents scaling on all samples showing a distribution shift and poses a concrete barrier to deploying models in all critical applications that require true generalization power. For instance, autonomous driving could face environments and circumstances not encountered during the training phase, caused by light, weather, background, and nearby objects dynamics. Indeed, disparate independent studies report how neural networks could easily fail without effective generalization capabilities, negatively affecting the behavior of the overall system \cite{dai2018dark,volk2019towards}. Similarly, another realistic example of a domain gap is training neural networks in simulation, which has become a standard procedure in the robotics research community. Recently, researchers have faced the Simulation-to-Reality (Sim2Real) gap problem, trying to effectively transfer Deep Neural Networks from virtual scenarios to the real world \cite{tobin2017domain,mozifian2020intervention}.

Domain Generalization (DG) aims at training models that generalize to out-of-distribution (OOD) data. The access to a set of source datasets provides a predictor with the ability to extract and learn general invariant patterns, which are, hypothetically, also recognizable in the target domain dataset \cite{blanchard2011generalizing,muandet2013domain}. As an extension of supervised learning, this approach aims to minimize empirical risk at training time to extrapolate an overall probability distribution from source datasets that enables accurate classification of OOD data.
In the last decade, aware of the tremendous impact of generalization on computer vision applications, the DG research community has tackled the problem with algorithms that aim to find invariances that hold with novel domains. Among the constellation of proposed approaches, we identify the principal broad strategies adopted for domain generalization in augmenting the source domain \cite{shankar2018generalizing,volpi2018generalizing}, aligning domain distributions \cite{ganin2016domain,sun2016deep,motiian2017unified,li2018domain, chen2023domain}, meta-learning \cite{balaji2018metareg,li2018learning,zhang2020adaptive}, self-supervised learning \cite{bucci2022self,albuquerque2020improving, rahman2020correlation}, and regularization strategies \cite{sagawa2019distributionally,huang2020self,shahtalebi2021sand,kim2021selfreg,segu2023batch}.

Although methodologies have given meaningful insights about the nature of DG over the years, only recent research contributions have proposed a rigorous testing benchmark to evaluate and compare the advantages provided by DG algorithms fairly. With \textsc{DomainBed} \cite{gulrajani2020search}, the results obtained by the most relevant solutions have been critically analyzed over DG datasets, unmasking the marginal positive or negative improvement obtained in most cases compared to naive empirical risk minimization (ERM). Nevertheless, the study has been carried out uniquely with ResNet50 \cite{He_2016_CVPR} as a feature extractor. Thus, new DG algorithms are still proposed overlooking a fundamental aspect of practical deep learning applications: the importance of the backbone. In past years, several competitive deep learning architectures, characterized by different types of feature extractors, have been proposed to solve classification tasks \cite{elharrouss2022backbones} on popular datasets such as ImageNet \cite{deng2009imagenet}. Classical backbones are based on convolutional layers: AlexNet \cite{krizhevsky2012imagenet} is a network based on a small set of convolutions and max-pooling layers in combination with ReLU activation. The VGG architecture \cite{simonyan2014very}, in its variations VGG-16 and VGG-19, further explores the convolution-pooling structure by stacking more layers and reaching a deeper design. ResNet \cite{he2016deep} first adopted a residual approach to help gradient flow with skip-connections, and it is still a widely adopted backbone for a variety of computer vision tasks.
Similarly to VGG, ResNet has been proposed in different fashions, with variable depth, such as ResNet18, ResNet34, and ResNet50. Other architectures, such as DenseNet \cite{huang2017densely} or InceptionNet \cite{szegedy2015going}, focus on different mechanisms, like dense connections or parallelization of convolutional layers with different kernel sizes. MobileNet \cite{howard2017mobilenets} and EfficientNet \cite{tan2019efficientnet} have been proposed to increase model efficiency, reaching competitive classification results with lightweight architectures and fewer parameters. More recently, self-attention-based models have reached state-of-the-art image classification performance, inspired by the Transformer\cite{vaswani2017attention} architecture first proposed for language modeling tasks. In particular, the Vision Transformer (ViT) \cite{dosovitskiy2021an} first adapted a Transformer encoder for vision tasks, while its training methodology has been refined by the Data Efficient ViT (DeiT)\cite{touvron2021training}; ConViT \cite{d2021convit} combines convolutions with self-attention, and LeViT\cite{graham2021levit} focuses on a pyramidal architecture of self-attention layers that progressively shrinks spatial dimensions. This rich literature landscape offers a wide choice for researchers when selecting feature extractors for visual applications. However, among the different computer vision tasks, the DG community has substantially neglected the generalization power of existing backbones, promoting sophisticated algorithms combined with outdated feature extractors such as ResNet18 or even AlexNet.

\begin{figure}[t]
  \centering
   \includegraphics[width=225pt]{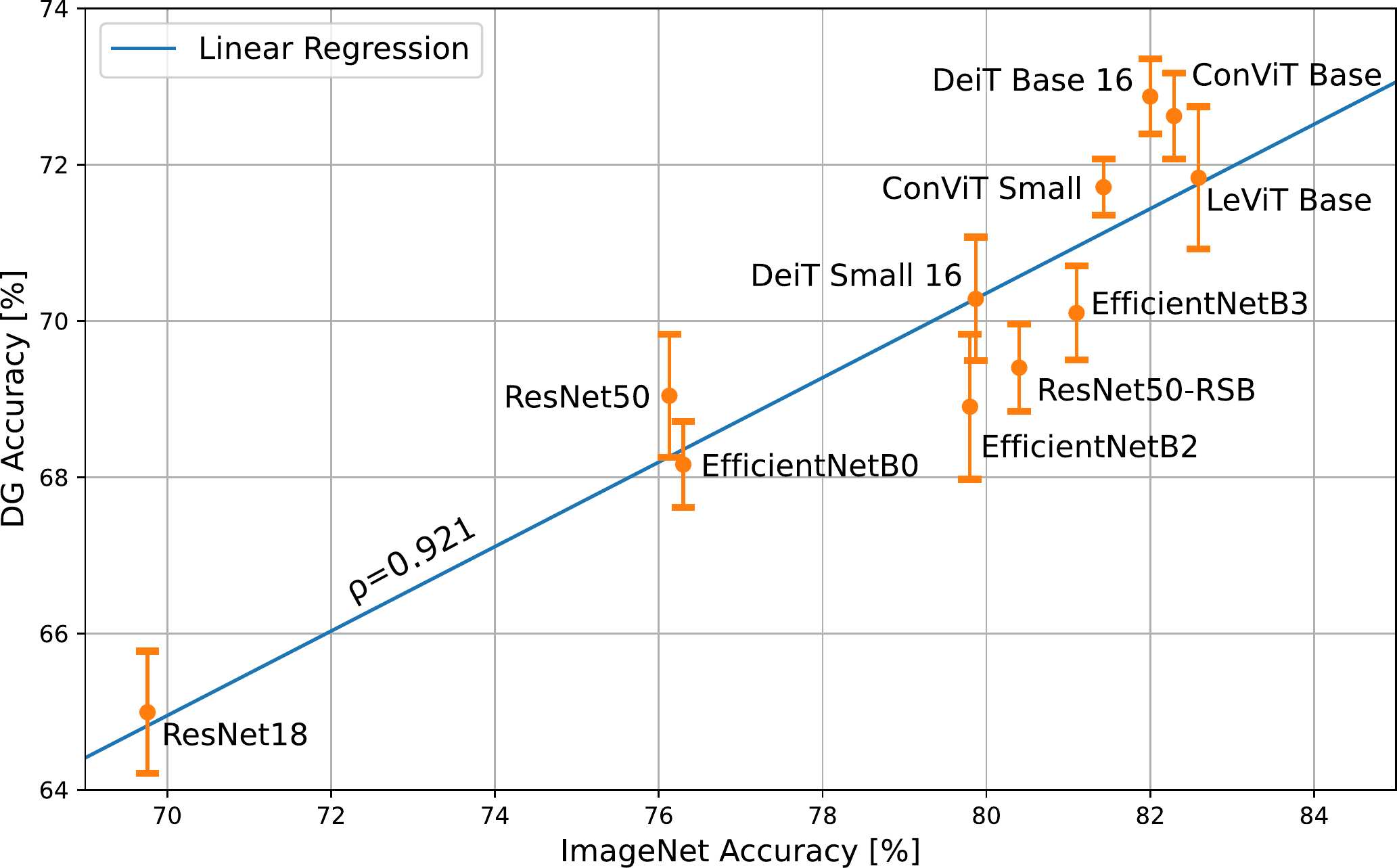}
   \caption{DG accuracy achieved by tested backbones compared with their performance on ImageNet, with error bars. Regardless of different architectures and priors, we find a strong linear correlation between the two metrics ($\rho=0.921$). In \cref{sec:baseline_benchmark}, we also compare DG accuracy with the number of parameters finding a much weaker correlation.}
   \label{fig:linear}
\end{figure}

In this paper, we claim that the domain gaps existing in realistic scenarios should be tackled starting from an accurate selection of the model architecture, which has an undeniable central role in most deep learning applications (\cref{fig:firstpage}). In particular, we conduct extensive experimentation on the principal DG datasets and assess a wide variety of backbone architectures, from novel vision transformers to standard convolutional models. Our results demonstrate an evident linear correlation between large-scale single-domain classification accuracy and domain generalization performance (\cref{fig:linear}). Moreover, we achieve state-of-the-art results in DG with naive ERM and simple data augmentation, remarking that, under fair testing conditions, the most promising algorithms presented so far give no substantial advantage.

We reinforce the experimentation with a visual analysis of the feature extractors. Using the t-SNE manifold learning technique \cite{van2008visualizing} on extracted features, we show that novel backbones map same-class samples closer in the feature space and outperform older architectures when trained in a DG framework. We propose a quantitative evaluation of this difference by fitting a k-NN classifier on the extracted features.

This study aims to promote a complete and meaningful approach to the domain generalization problem, avoiding isolated research efforts on DG algorithms and encouraging contributions that target the overall maximization of model generalization. Evidence in the literature shows that researchers from disparate application fields could significantly benefit from a shift of the DG paradigm towards realistic circumstances. For instance, data augmentation can automatically be exploited to generate a vast collection of artificial source domains. Domain Randomization fully exploits this principle \cite{tobin2017domain}, demonstrating its effectiveness in training agents in simulation for controlling manipulators accomplishing visual tasks \cite{tobin2018domain} and autonomous racing drones \cite{loquercio2019deep}. That is further concrete proof that the success of domain generalization in real-world applications relies on simple ERM techniques, which offer an easy implementation together with a robust generalization boost.

The main contributions of this work can be therefore summarized as follows:
\begin{itemize}
    \item We propose an extensive evaluation of backbones for domain generalization, showing remarkable improvements compared to literature results. We empirically find a linear correlation between large-scale single-domain classification accuracy and domain generalization performance (\cref{fig:linear}).
    \item We prove that adopting DG algorithms does not provide the expected generalization boost compared to naive ERM when using state-of-the-art feature extractors.
    \item We enrich the conducted experiments with an introspective study of the backbones, comparing the feature representations before and after the DG fine-tuning.
\end{itemize}

As an outcome of this work, we release \textsc{Back-to-Bones}\footnote{\url{https://github.com/PIC4SeR/Back-to-Bones}}, a testbed to encourage the deep learning community to evaluate and compare the domain generalization performance of newly proposed backbones.

The rest of the paper is organized as follows. In \cref{sec:dg_theory}, we briefly frame the DG theoretical background and introduce our backbone definition. In \cref{sec:back-to-bones}, we introduce our research outcome, describing the conducted approach and the criteria which guided the choice of backbones, model selection, hyperparameter optimization, and overall experimental framework; then, we report numerical results in conjunction with a visual introspection of the representations learned by the most relevant backbones under investigation. \cref{sec:additional_consideration} discusses additional considerations about transformer-based backbones generalization as well as on baseline selection in previous works. Finally, in \cref{sec:conclusion}, we present our conclusive remarks and suggestions for future works on DG.

\section{The Domain Generalization Framework}
\label{sec:dg_theory}

In this section, we first define necessary notations and concepts to frame the problem of domain generalization and empirical risk minimization. Secondly, we introduce a formal definition of a backbone and its constituents.

\begin{description}[leftmargin=0pt]

\item[Problem Definition] Given the input random variable \(X\) with values \(x \in \mathcal{X}\), and the target random variable \(Y\) with values \(y \in \mathcal{Y}\), the definition of \textit{domain} is associated with the joint probability distribution \(P_{XY}\), or \(P(X,Y)\), over \(\mathcal{X}\)x\(\mathcal{Y}\). Supervised learning aims to train a classifier \(f:\mathcal{X} \to \mathcal{Y}\) exploiting $N$ available labeled examples of a dataset   \(D = {(x_i, y_i)}^N_{i=1}\) that are identically and independently distributed and sampled according to \(P_{XY}\). The goal of the training process is to minimize the \textit{empirical risk} associated with a loss function \(l:\mathcal{Y}\times\mathcal{Y} \to [0,+\infty)\), 
\begin{equation}
    R_{\text{emp}}(f)={\frac {1}{N}}\sum _{i=1}^{N}l(f(x_{i}),y_{i})
\end{equation}

\noindent by learning the classifier \(f\). The dataset D is the only available source of knowledge to learn \(P_{XY}\). We refer to this basic learning method as empirical risk minimization \cite{vapnik1999overview}. 

In domain generalization, a set of different \(K\) source domains \(\mathcal{S}=(S_k)^K_{k=1}\) is used to learn a classifier \(f\) that aims at generalizing well on an unknown target domain \(T\). Each source domain is associated with its joint probability distribution \(P_{XY}^k\), whereas \(P_{XY}^\mathcal{S}\) indicates the overall source distribution learned by the classifier \cite{zhou2021domain}. Indeed, DG aims to enable the classifier to predict well on out-of-distribution data, namely on the target domain distribution \(P_{XY}^T\), by learning an overall domain invariant distribution from the source domains seen during training.

\item[Backbone Definition] We define a backbone $\B=f(\A,\T_{B},\D)$ as a function of three elements: the model architecture $\A$, the training procedure $\T_{B}$ (including optimization, regularization, and data augmentation), and the training data $\D$. Consequently, all three factors introduce a certain degree of variability to the domain generalization accuracy:
\begin{center}
    DG\textsubscript{accuracy} ($\mathcal{S}$, $T$) = $g(\B, \T_{DG}, \N_{exp})$
\end{center}
where $\T_{DG}$ is the adopted DG training procedure and $\N_{exp}$ is the experimentation noise. $\T_{DG}$ usually includes a dedicated algorithm to cope with domain shifts. $\N_{exp}$ comprehends a systematic error due to the adopted model selection strategy and a random component caused by the stochasticity in the training process.
\end{description}

\section{Back-to-Bones}
\label{sec:back-to-bones}
We set up our experimental benchmark to run a detailed analysis of the role of feature extractors in domain generalization. Besides choosing architectures, datasets, and DG algorithms to evaluate, particular attention is given to model selection strategy and statistical interpretation to obtain a fair and accurate benchmark. In the following subsections, we provide details on our experimental setup.

\begin{table}[ht]
\centering
\resizebox{350pt}{!}{
\begin{tabular}{@{}cccccc|c|c@{}}
\toprule
{\textbf{\centered{Backbone}}} &
  {\textbf{PACS}} &
  {\textbf{VLCS}} &
  {\textbf{Office-Home}} &
  {\textbf{Terra Incognita} } &
  {\textbf{Average}} & 
  {\textbf{ImageNet}} & 
  {\textbf{Parameters}}\\ \midrule
{ResNet18} &
  {80.51 ± 0.29} &
  {74.64 ± 0.61} &
  {63.87 ± 0.36} &
  {40.93 ± 1.85} &
  {64.99 ± 0.78} &
  {69.76} &
  {11.69M} \\
{ResNet50 \cite{gulrajani2020search}} &
  85.50  ± 0.20 &
  77.50 ± 0.40 &
  66.50  ± 0.30 &
  46.10  ± 1.80 &
  68.90  ± 0.68 &
  {76.13}  &
  {25.56M} \\
{ResNet50} &
  {83.85 ± 0.77} &
  {76.21 ± 1.20} &
  {68.79 ± 0.21} &
  {47.32 ± 0.97} &
  {69.04 ± 0.79} &
  {76.13} &
  {25.56M} \\ 
{ResNet50 A1} &
  {84.52 ± 0.68} &
  {78.37 ± 0.56} &
  {72.47 ± 0.13} &
  {42.23 ± 0.87} &
  {69.40 ± 0.56} &
  {80.40} &
  {25.56M} \\
 \midrule
{EfficientNetB0} &
  {85.46 ± 0.65} &
  {75.16 ± 0.34} &
  {67.27 ± 0.27} &
  {44.76 ± 0.94} &
  {68.16 ± 0.55}&
  {76.30} &
  {5.29M}  \\
{EfficientNetB2} &
  {87.02 ± 1.37} &
  {75.44 ± 0.20} &
  {69.35 ± 0.24} &
  {43.80 ± 1.90} &
  {68.90 ± 0.93} &
  {79.80} &
  {9.11M} \\ 
{EfficientNetB3} &
  {86.71 ± 0.30} &
  {78.14 ± 0.18} &
  {69.84 ± 0.08} &
  {45.70 ± 1.84} &
  {70.10 ± 0.60} &
  {81.10} &
  {12.23M} \\ \midrule
{DeiT Small 16} &
  {86.22 ± 1.33} &
  {79.47 ± 0.41} &
  {72.03 ± 0.33} &
  {43.40 ± 1.08} &
  {70.28 ± 0.79}&
  {79.87} &
  {22.05M}  \\
{DeiT Base 16} &
  {\bf{88.10 ± 0.48}} &
  {79.80 ± 0.32} &
  {76.35 ± 0.36} &
  {47.22 ± 0.75} &
  {72.87 ± 0.48}&
  {82.00} &
  {86.57M}  \\
{ConViT Small} &
  {87.10 ± 0.33} &
  {80.00 ± 0.34} &
  {73.90 ± 0.17} &
  {45.83 ± 0.61} &
  {71.71 ± 0.36} &
  {81.43} &
  {27.78M} \\
{ConViT Base} &
  {87.27 ± 0.40} &
  {\bf{80.31 ± 0.67}} &
  {76.51 ± 0.25} &
  {46.37 ± 0.89} &
  {72.62 ± 0.55}&
  {82.29} &
  {86.54M}  \\
{LeViT Base} &
  {87.55 ± 1.50} &
  {78.91 ± 0.50} &
  {75.16 ± 0.13} &
  {45.68 ± 1.50} &
  {71.83 ± 0.91} &
  {82.59} &
  {39.13M} \\ \midrule
{ViT Small 16*} &
  83.59 ± 0.43 &
  79.96 ± 0.60 &
  77.25 ± 0.33 &
  44.12 ± 1.07 &
  71.23 ± 0.61 &
  {81.40} &
  {22.05M} \\
{ViT Base 32*} &
  {84.00 ± 1.17} &
  {78.46 ± 0.64} &
  {76.84 ± 0.17} &
  {36.71 ± 2.07} &
  {69.00 ± 1.01} &
  {80.72} &
  {88.22M} \\
{ViT Base 16*} &
  \bf{88.48 ± 1.22} &
  \bf{80.05 ± 0.15} &
  \bf{81.47 ± 0.21} &
  \bf{49.77 ± 1.28} &
  \bf{74.94 ± 0.72} &
  {84.53} &
  86.57M \\ \bottomrule
\end{tabular}}
\caption{Baselines comparison of different backbones for DG. We report the average accuracy over three runs and the associated standard deviation for each model. We include the results achieved by \textsc{DomainBed} with ResNet50 for reference. The models marked with * are pretrained on Imagenet21K instead of ImageNet1K. The rightmost column indicates the accuracy of the networks on ImageNet1K. In \cref{appsec:additional-benchmark}, we report in detail the results obtained for all the domains.}
\label{tab:baselines}
\end{table}

\begin{description}[leftmargin=0pt]
\item[Backbones] To be consistent with previous works, we include ResNet18 and ResNet50 \cite{He_2016_CVPR} in the benchmark and compare them with some of the most successful architectures proposed in recent image classification research. We also consider the latest ResNet50 A1 \cite{wightman2021resnet}, trained using the most recent practices in optimization and data augmentation and reaching a remarkable 80.4\% top-1 accuracy on Imagenet1K. We include different sizes for each network to glimpse the effects of model dimension on DG accuracy. EfficientNet \cite{pmlr-v97-tan19a} demonstrated that systematical model scaling and dimension balancing yield remarkable results with fewer parameters. For this reason, we select three network versions, namely B0, B2, and B3. Finally, transformers \cite{vaswani2017attention} recently revolutionized deep learning by proving the effectiveness of self-attention for feature extraction; hence four transformer-based architectures are included in the comparison. In particular, we choose DeiT (Small and Base) \cite{touvron2021training}, ConViT \cite{d2021convit} (both in its Small and Base configurations), and LeViT Base \cite{graham2021levit}. To provide further insights on the effect of additional pretraining data besides standard ImageNet \cite{deng2009imagenet}, we also include Vision Transformer (ViT) \cite{dosovitskiy2021an} trained on ImageNet21K in its Small and Base versions. Regarding ViT Base, a configuration with a 32x32 patch size has been added to the standard 16x16 format to test the impact of patch size on DG. Further information on architectural details can be found in the cited papers. We report the number of parameters for each model in the last column of \cref{tab:baselines}.

\item[Datasets] Among the various datasets created explicitly for DG in the last years, we use four of the most widely adopted ones for our primary experimentation. VLCS \cite{Fang_2013_ICCV} considers four previous classification datasets as domains, while PACS \cite{Li_2017_ICCV} and Office-Home \cite{Venkateswara_2017_CVPR} focus more on style shifts (e.g. from photos to cartoons, sketches, and paintings). Terra Incognita \cite{Beery_2018_ECCV} comprehends several animal photos taken with camera traps placed in different locations by day and night. To those, we add DomainNet \cite{Peng_2019_ICCV}, a bigger and more recent dataset that contains six domains divided by style and 345 classes. We use it to further stress the generalization capability of the best-performing backbones in the presence of more transfer learning data and fewer samples per class. We omit Rotated MNIST \cite{ghifary2015domain} and Colored MNIST \cite{arjovsky2019invariant} since we consider them too distant from any practical application. Moreover, from our perspective, simple rotation and colorization do not constitute actual domain shifts.

\item[DG Algorithms] We choose some of the most promising DG algorithms in recent research, particularly considering their performance on \textsc{DomainBed} \cite{gulrajani2020search}. Moreover, we select them to explore different approaches to the DG problem. CORAL \cite{sun2016deep} and MMD \cite{li2018domain}, indeed, focus on aligning the extracted features through second-order statistics (covariance). On the other hand, Mixup \cite{yan2020improve} works directly on input images, interpolating samples from different domains and considering the loss coming from both precursors. RSC \cite{huang2020self}, instead, introduces a heuristic that discards dominant features in the label determination, stimulating the model to rely on weaker data correlations. CausIRL \cite{chevalley2022invariant} (used in combination with MMD or CORAL) builds from a causal analysis of generalization enforcing soft domain invariance to interventions on the source domain. CAD \cite{ruan2022optimal} introduces a contrastive adversarial domain bottleneck to guarantee convergence to target domains that preserve the Bayes predictor. ADDG \cite{meng2022attention} exploits a double mechanism (Intra-model and Inter-model) to provide attention diversification between features and suppress domain-related attention.

\item[Data Augmentation] Many research works prove that data augmentation plays a fundamental role in DG, as it can partially compensate for certain domain shifts \cite{volpi2018generalizing}. That is particularly true in the presence of style changes, as popular data augmentation strategies involve the alteration of saturation, hue, and contrast. Since the effect of data augmentation on DG has already been investigated, in this paper, we use a standard setup to keep the focus on backbones.
The de-facto standard augmentation strategy for DG, which we use in our benchmark, includes random cropping keeping at least 80\% of the original image, horizontal flipping with 50\% probability, image grayscaling with 10\% chance, and random changes in color brightness, contrast, saturation, and hue, with a maximum of 40\%. Since all the models are pretrained on ImageNet1K or ImageNet21K, input images are further normalized according to the mean and standard deviation of that datasets.

\item[Model Selection] To assess the DG capability of the considered pretrained networks, we fine-tune each of them on a set of \(K\) source domains \(\mathcal{S}\) and test them on a target domain \(T\). As pointed out by \cite{gulrajani2020search}, ``a domain generalization algorithm should be responsible for specifying a model selection method" and avoid improper comparisons between results obtained adopting different selection methods. In total agreement with their recommendations, we use the \textit{training-domain validation set} strategy, which picks the model maximizing the accuracy on a validation split of the training set (in our case 10\%, uniform across domains) at the end of each epoch. This selection method assumes that the average distribution of source domains is similar to that of the target domain on which the best model is tested.

\item[Hyperparameter Search] We conduct a random search for each backbone and dataset to determine the optimal training hyperparameters for the baselines. We define a range of values for continuous arguments and a set of choices for discrete ones, running approximately 32 iterations for each search and selecting the best combination via the previously defined model selection strategy. The learning rate is bounded in the range $[10^{-6},10^{-2}]$, choosing its scheduler among step (90\% reduction after 80\% of the epochs), exponential (with a decay in the range $[0.9,1)$), and cosine annealing. The batch size and the number of training epochs are the same for all the experimentation, fixing their values at 32 and 30, respectively. Finally, we use cross-entropy loss and select the optimizer among SGD (with a momentum of $0.9$) and Adam, keeping the weight decay to $5\cdot10^{-4}$.

\item[Experimental Framework] Our benchmarks are developed in Python 3 using the deep learning framework PyTorch. As the experimentation applies transfer learning to pretrained models, we use existing implementations of the considered backbones. Only the classification head is changed, adapting the network to the different number of classes. In particular, standard ResNets are taken from the PyTorch library \textit{torchvision}\footnote{pytorch.org/vision/stable/models}, EfficientNets from \textit{EfficientNet-PyTorch}\footnote{github.com/lukemelas/EfficientNet-PyTorch}, transformers and ResNet50 A1 from \textit{timm}\footnote{github.com/rwightman/pytorch-image-models}. The implementations of DG algorithms are taken from \textsc{DomainBed}\footnote{github.com/facebookresearch/DomainBed} and adapted to work with the architectures under test.

We repeat each training three times with different and randomly generated seeds to give more statistical information about accuracy results. In this way, both hyperparameter search and benchmarks cannot take advantage of the repeatability of trials, as data splitting, augmentation, and weight initialization change from one iteration to the next. Therefore, each of the results of our benchmark is reported as the mean over three repetitions, along with its standard deviation.
\end{description}

\subsection{Baseline Benchmark}
\label{sec:baseline_benchmark}
The first analysis of our work consists of a precise and fair benchmark of the DG capabilities of recent deep learning architectures for image classification, trying to determine what solutions work best and, possibly, why. Every pretrained backbone, after a hyperparameter search, is trained following the standard DG \textit{leave-one-domain-out} procedure using the previously described model selection strategy. Our benchmark results are reported in \cref{tab:baselines} as the mean and standard deviation over three iterations. 

\begin{table}[t]
\centering
\resizebox{200pt}{!}{
\begin{tabular}{@{}cccccccc@{}}
\toprule
\textbf{Backbone} & \textbf{C} & \textbf{I} & \textbf{P} & \textbf{Q} & \textbf{R} & \textbf{S} & \textbf{Avg} \\ \midrule
ResNet50\cite{gulrajani2020search}     & 58.1          & 18.8          & 46.7          & 12.2          & 59.6 & 49.8 & 40.9 \\
DeiT Base 16 & 69.1          & 25.0          & 55.8          & 17.1          & 69.3  & 57.0  & 48.9  \\
ConViT Base  & 69.5          & 24.3          & 55.7          & \textbf{17.7} & 69.3 & 57.0  & 48.9  \\
ViT Base 16* & \textbf{74.9} & \textbf{28.9} & \textbf{60.8} & 17.5          & \textbf{77.3}  & \textbf{61.8}  & \textbf{53.5}  \\ \bottomrule
\end{tabular}
}
\caption{Baseline comparison of a selection of the best backbones on DomainNet (\textit{Clipart}, \textit{Infograph}, \textit{Painting}, \textit{Quickdraw}, \textit{Real}, and \textit{Sketch} domains). We include the results achieved by \textsc{DomainBed} with ResNet50 for reference. The model marked with * is pretrained on Imagenet21K instead of ImageNet1K.}
\label{tab:domainnet}
\end{table}

Firstly, our benchmark highlights a strong correlation between DG accuracy and ImageNet performance. As depicted in \cref{fig:linear}, we find a direct proportionality between the two metrics (excluding the ViT models due to their different pretraining). Applying linear least-square regression, we obtain a Pearson correlation coefficient ($\rho$) of 0.921. Indeed, a quick look at the results is sufficient to notice how newer and more performing backbones tend to achieve a higher DG accuracy on nearly all the datasets. That is primarily true for different sizes of the same architecture. ResNet50 reaches better results than ResNet18 for all the datasets, and the same happens for EfficientNet, ConViT, and ViT variants. For ResNet50, we also compare our results with those obtained by \textsc{DomainBed} and find comparable values. ResNet50 A1 proves to benefit from its stronger pretraining, largely improving the accuracy obtained by the standard model on VLCS and Office-Home. However, Terra Incognita  seems to penalize the network with its peculiar light conditions, resulting in a slight overall enhancement.
Regarding different architectures, EfficientNetB2 performs very similarly to ResNet50 while the B3 version gains an additional 1\% on them. Transformer-based models bring further improvements by exploiting their self-attention-based feature extraction, even in the case of  DeiT Small and ConViT Small. In particular, they strongly outperform EfficientNet on OfficeHome by over 4\%, while Terra Incognita is the only dataset without any significant progress. That is probably due to the peculiarity of the domains, comprehending many night shots that can be challenging even for humans and rewarding less an effective ImageNet pretraining. Among other transformers, DeiT Base 16 and ConViT Base prove to be the best, the latter being slightly more performing. Finally, the three ViT models show that pretraining on a more significant amount of data improves generalization. However, only ViT Base 16 registers a considerable step forward, suggesting that the abundance of data is fully exploited only by larger models. Nonetheless, ConViT Small performs similarly to the same-sized ViT Small 16, while larger patches demonstrate to degrade the accuracy of ViT Base 32.
In conclusion, our results show how better DG comes from the union of a good feature extractor architecture and an optimal pretraining, as none of the two is sufficient alone. In \cref{sec:additional_consideration}, we further discuss the generalization capability of transformers. We stress the importance of adopting a good model selection strategy by comparing our ResNet18 baseline with various recent results obtained using the same backbone.

\begin{figure}[t]
    \centering
    \includegraphics[width=225pt]{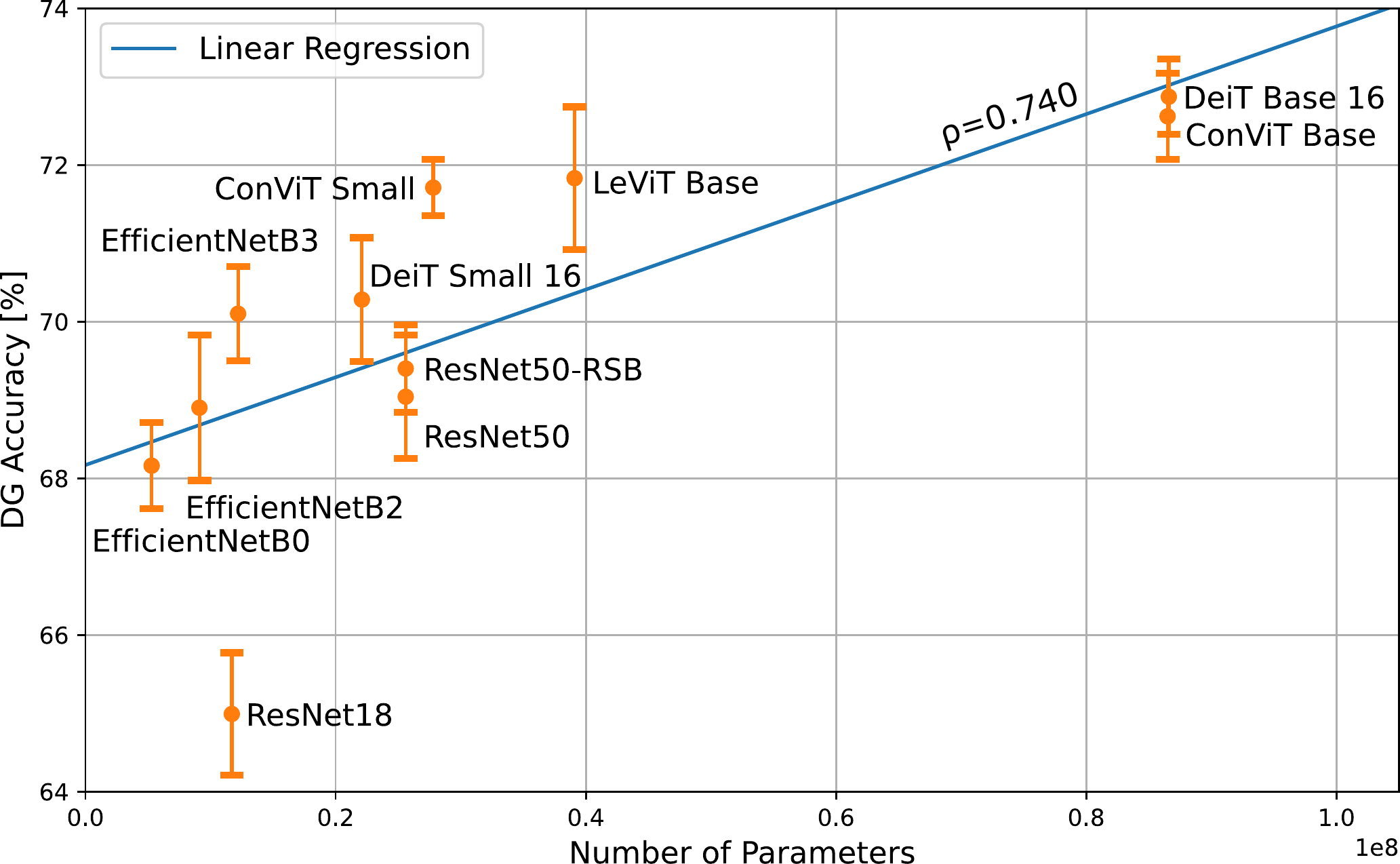}
    \caption{DG accuracy achieved by tested backbones compared with their number of parameters, with error bars. We find a much weaker correlation between the two metrics ($\rho=0.740$) than the one reported in \cref{fig:linear}.}
    \label{fig:nonlinear}
\end{figure}

As an additional comparison, we plot the achieved DG accuracy compared to the number of parameters of the backbones (\cref{fig:nonlinear}). Contrary to the graph of \cref{fig:linear}, in this case, the correlation between model dimension and generalization is much less marked, with a Pearson correlation coefficient ($\rho$) of 0.740. This confirms the central role of model architecture in DG tasks and our idea of backbone as the union of architecture, training procedure, and data.

Finally, we conduct an additional benchmark on the DomainNet dataset. Although representing a significant challenge for large-scale generalization, we choose to include DomainNet only in this second stage of the study due to its demanding computational nature and strong class unbalancing. Indeed, our main intention is to promote a practical and accessible benchmark that aims to become a widespread reference for DG. We select only the best three models from the previous tests for this one (DeiT Base 16, ConViT Base, and ViT Base 16). In \cref{tab:domainnet}, we report the results achieved on each test domain, including those obtained by \textsc{DomainBed} on ResNet50 for reference. It is well evident that the feature extraction capabilities of modern backbones bring substantial improvement in all the domains, with an average increase in DG accuracy up to 12.6\%. Moreover, ViT further enhances the results by exploiting its stronger pretraining. 

\subsection{Model Introspection}
\label{sec:model_introspections}
After assessing the DG performance of different backbones, we propose a series of insights on how different architectures leverage training data to create their inner representation. First, we investigate the benefits of ImageNet pretraining for DG with a k-NN classifier, comparing ResNet50 and the best models from our benchmark. Then, we apply t-SNE \cite{van2008visualizing} on the same extracted features to visualize how close same-class and same-domain samples are and the effect of fine-tuning on DG datasets. Finally, we inspect the attention maps of one of the transformer-based models to have a qualitative insight on the region of the images it focuses on.

\begin{figure}[ht]
   \centering
   \includegraphics[width=250pt]{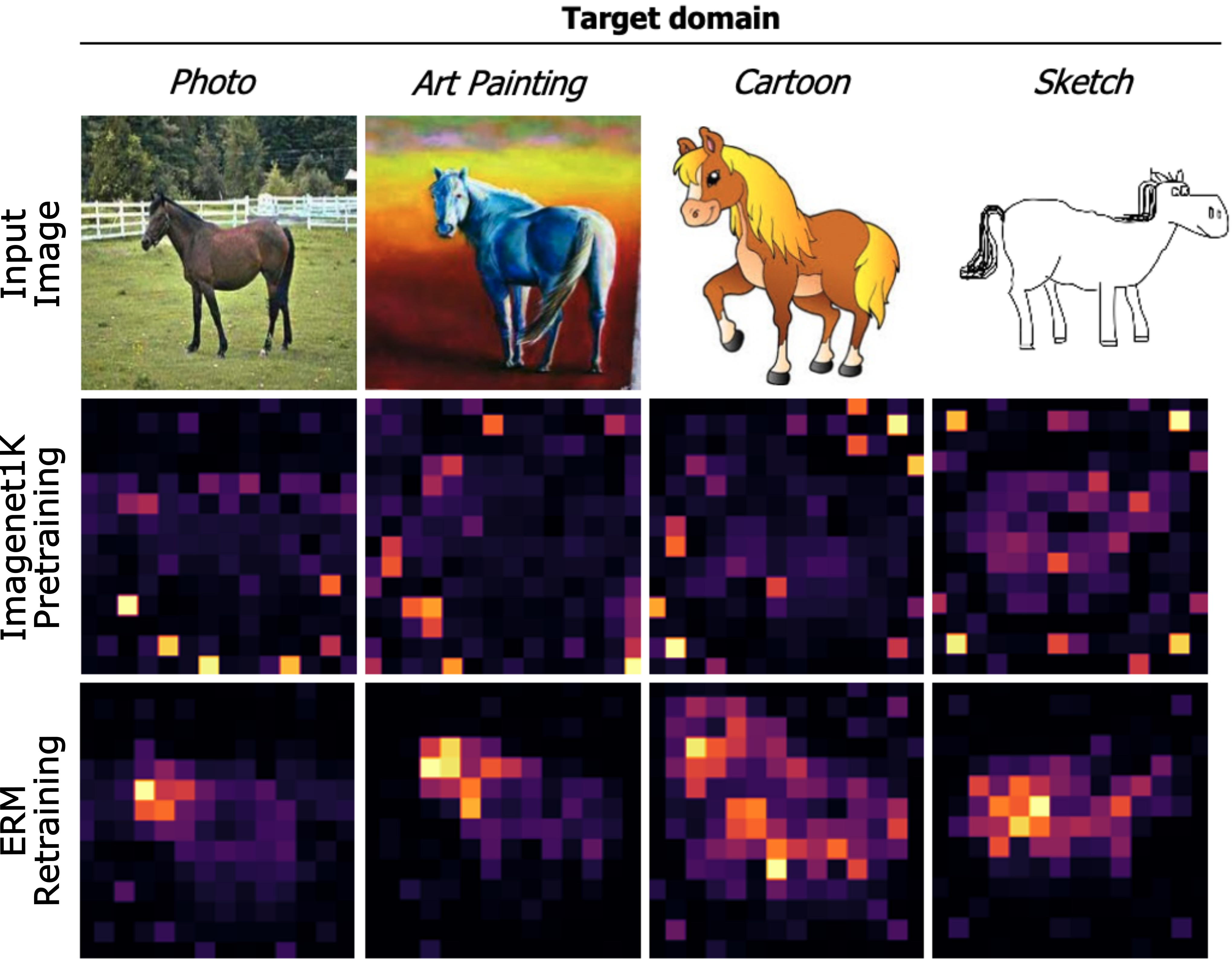}
   \caption{DeiT Base attention maps when using the {\fontfamily{qcr}\selectfont [CLS]} token as a query for the different heads in the last layer. We select the same head for all examples. ERM encourages the backbone to focus on domain-invariant features, highly mitigating pretraining noise.}
   \label{fig:attention_visualization}
\end{figure}

\begin{figure}[ht]
    \begin{subfigure}{\linewidth}
        \centering
        \includegraphics[width=0.6\columnwidth, trim={10px 10px 10px 10px}, clip]
        {legend_drawio_hor.png}
        \label{fig:legend}
    \end{subfigure}
    \centering
  \begin{subfigure}{0.4\linewidth}
    \centering
    \includegraphics[width=\columnwidth, trim={40px 45px 30px 40px}, clip]
                    {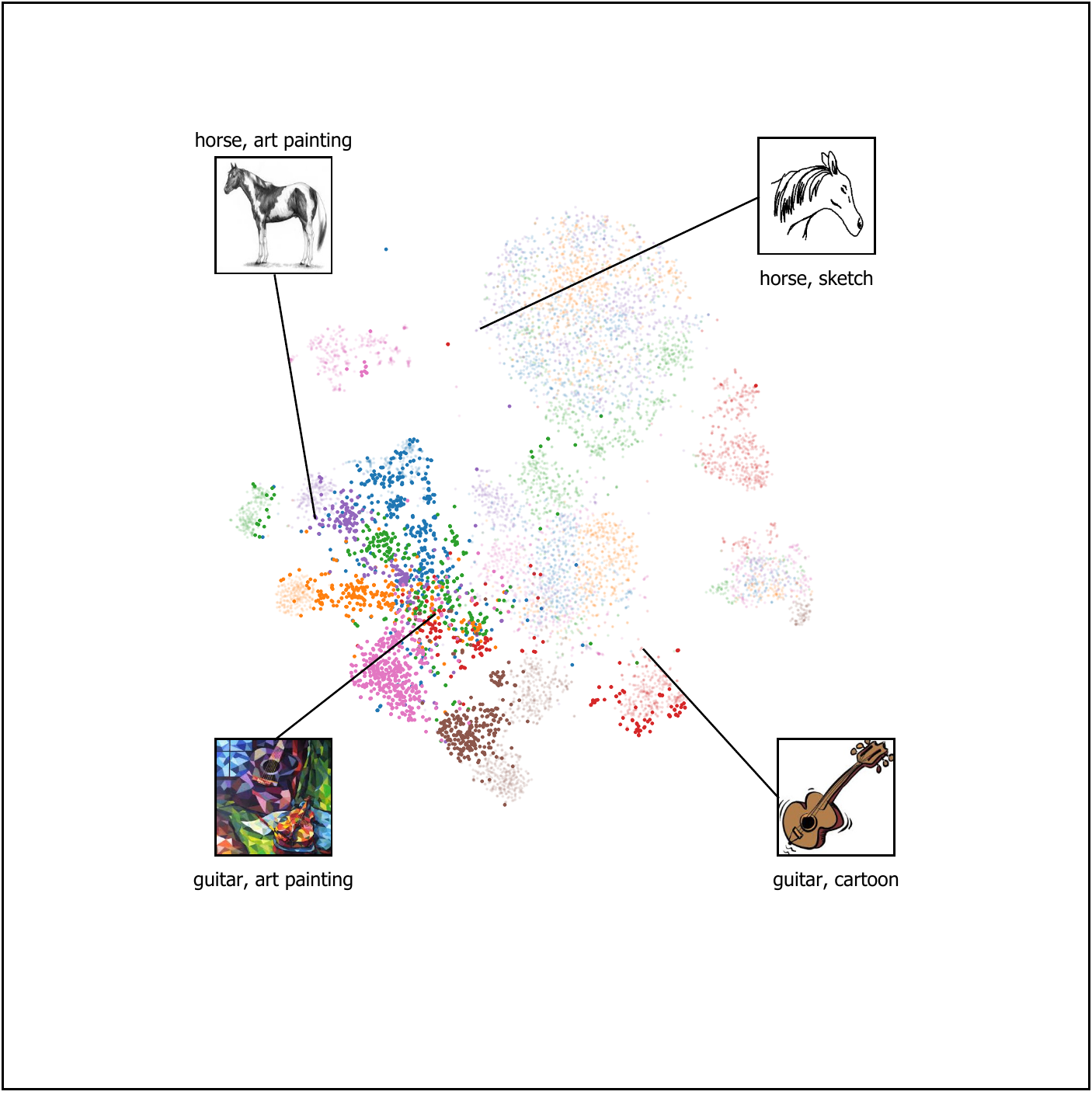}
    \caption{ResNet50 (ImageNet1K)}
    \label{fig:resnet50_imagenet}
  \end{subfigure}
  \begin{subfigure}{0.4\linewidth}
    \centering
    \includegraphics[width=\columnwidth, trim={40 45px 30px 40px}, clip]
                    {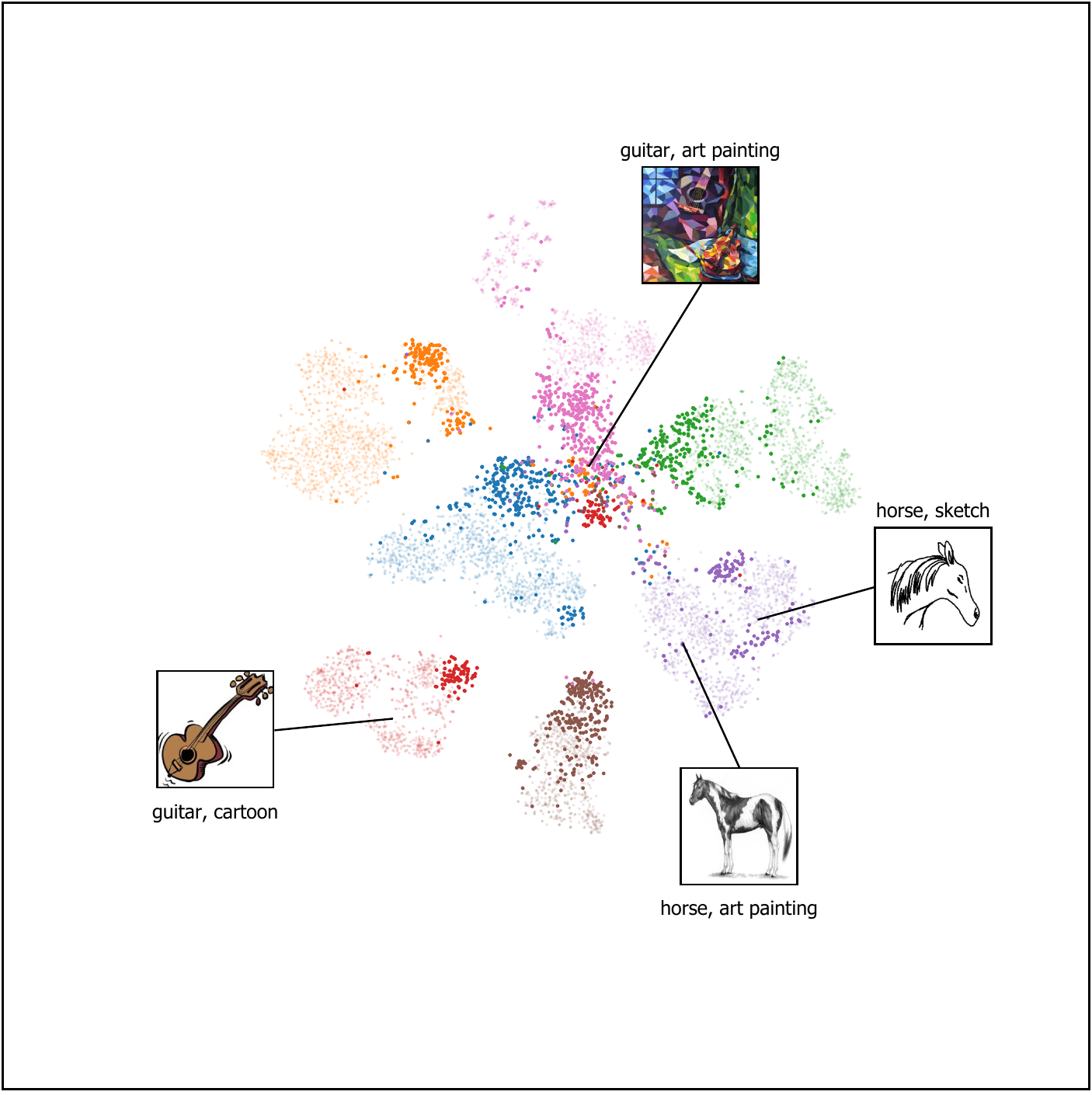}
    \caption{ResNet50 ($\mathcal{S}$: [\textit{P}, \textit{C}, \textit{S}]
                       $\rightarrow$ $T$: \textit{A})}
    \label{fig:resnet50_art_painting}
  \end{subfigure}
  \begin{subfigure}{0.4\linewidth}
    \centering
    \includegraphics[width=\columnwidth, trim={40 45px 30px 40px}, clip]
                    {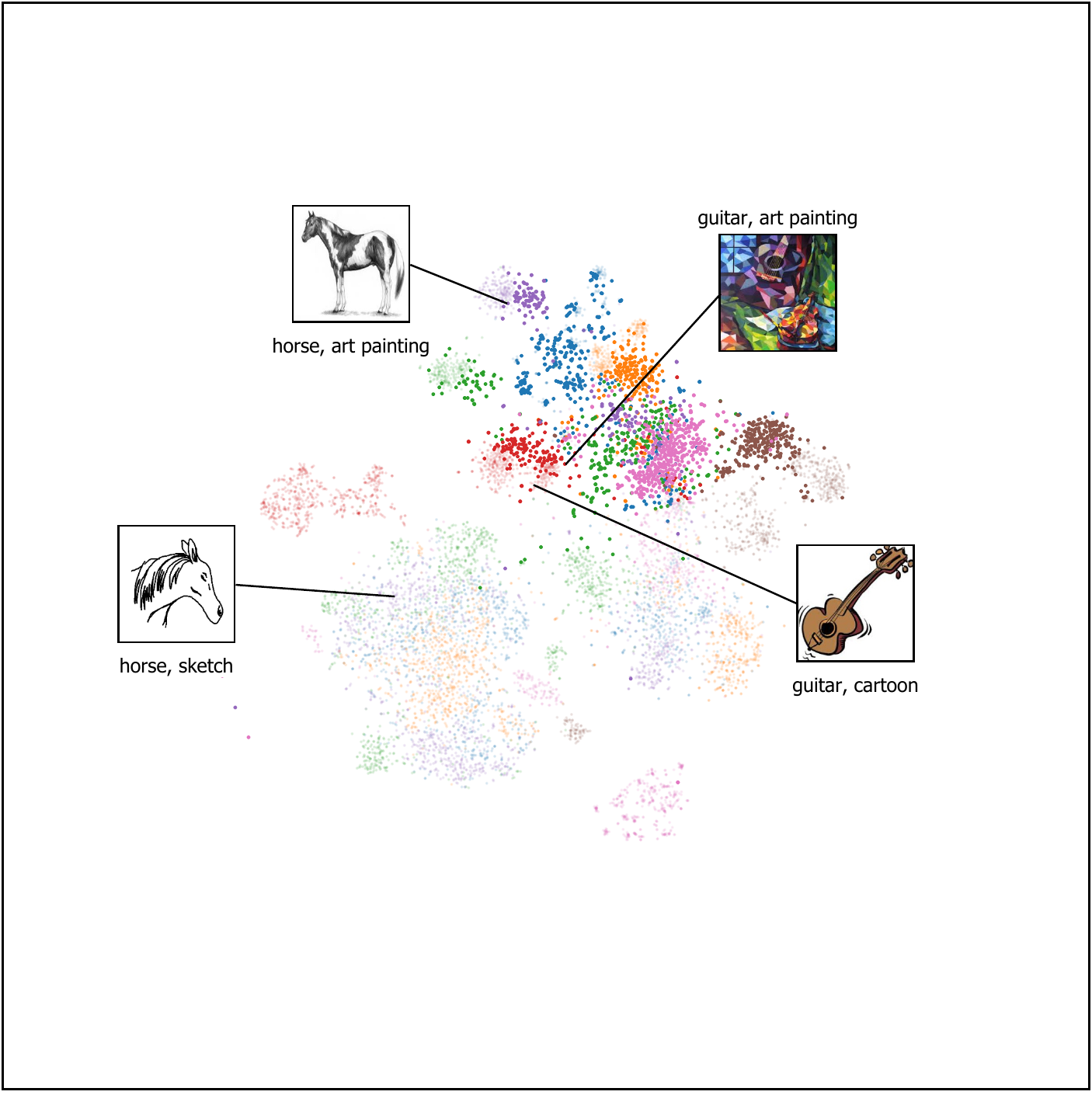}
    \caption{ConViT Base (ImageNet1K)}
    \label{fig:convit_base_imagenet}
  \end{subfigure}
  \begin{subfigure}{0.4\linewidth}
    \centering
    \includegraphics[width=\columnwidth, trim={40 45px 30px 40px}, clip]
                    {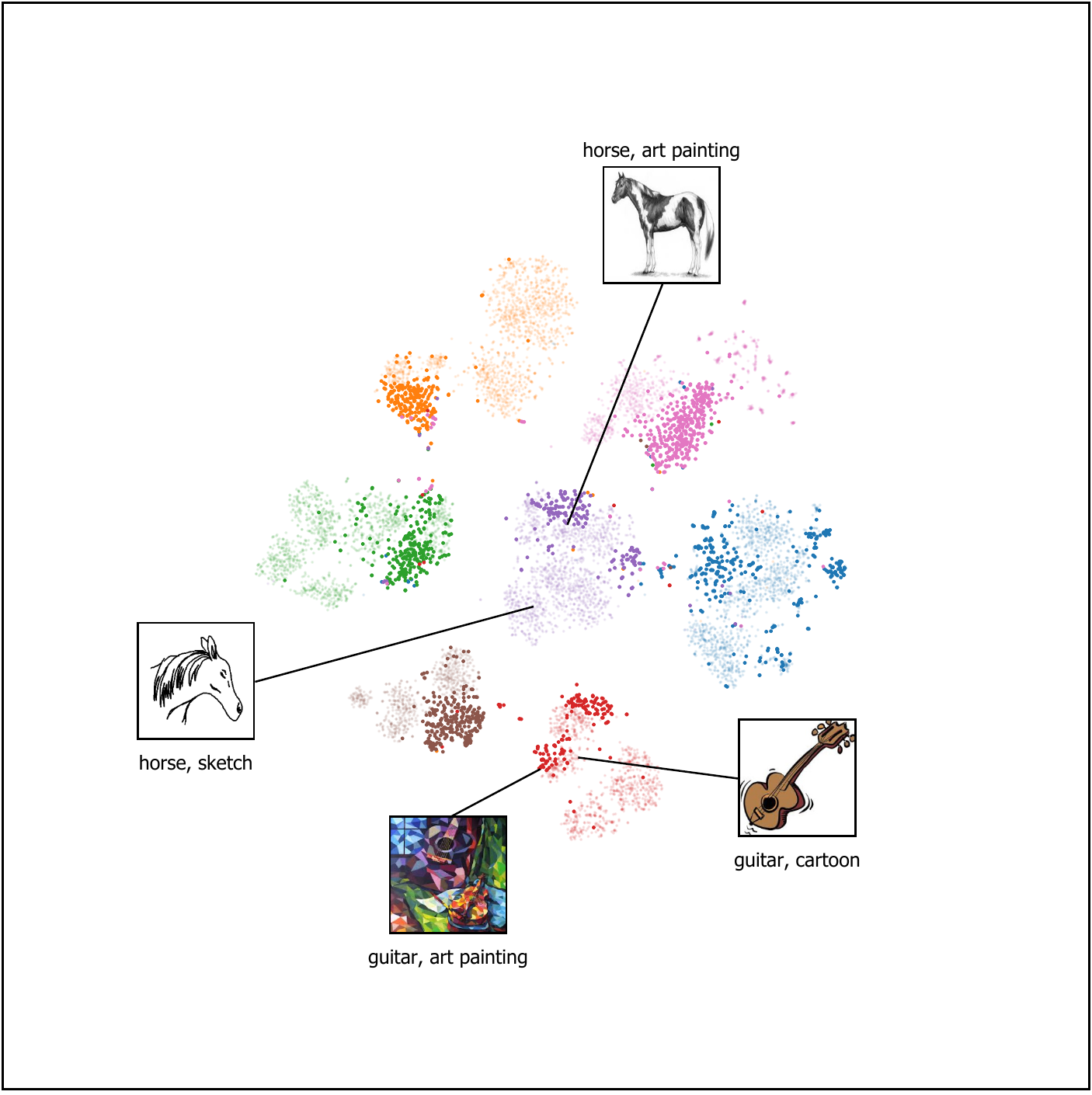}
    \caption{ConViT Base ($\mathcal{S}$: [\textit{P}, \textit{C}, \textit{S}]
                          $\rightarrow$ $T$: \textit{A})}
    \label{fig:convit_base_art_painting}
  \end{subfigure}
  \caption{Backbone features visualization with t-SNE on PACS (Photo (\textit{P}), Art Painting (\textit{A}), Cartoon (\textit{C}) and Sketch (\textit{S}) domains). Target domain samples are highlighted. Some image examples from different domains and classes are visualized for better interpretability. After the fine-tuning, the ConViT Base architecture achieves a better class separation than ResNet50, clustering together same-class samples of different domains.}
  \label{fig:tSNE}
\end{figure}

\begin{description}[leftmargin=0pt]
\item[K-NN Evaluation]
Firstly, we take ResNet50 and the best-performing models from our benchmark and evaluate their ability to tackle DG without fine-tuning. To do so, we use ImageNet weights to extract features from training domains and a k-NN (with $k=5$) to fit that data. Then, we use test-domain images for the evaluation. To have a fair comparison with our benchmark, we use the same amount of training data leaving out 10\% of samples from source domains. The results in \cref{tab:knn} show an overall difference of about 5\% between ResNet50 and transformer-based models pretrained on ImageNet1k.
This outcome is consistent with the generalization boost achieved in the standard DG framework (\cref{tab:baselines}), although k-NN results tend to oscillate among different datasets.
On the same trend, ViT Base 16 gains an additional 10\% average accuracy, thanks to its pretraining on the larger ImageNet21K dataset. This outcome suggests that learning a wider overall source distribution \(P_{XY}^\mathcal{S}\) is always needed to tackle a substantial domain gap effectively. That pretraining alone does not guarantee the ability to extract domain-invariant features.

\begin{table}[t]
\centering
\resizebox{200pt}{!}{
\begin{tabular}{@{}cccccc@{}}
\toprule
\textbf{Backbone} & \textbf{PACS}         & \textbf{VLCS}         & \textbf{Office-Home}  & \textbf{TerraInc.} & \textbf{Avg}      \\ \midrule
ResNet50 & 56.04 & 69.57 & 56.26 & 14.75 & 49.16 \\
DeiT Base 16 & 56.27 & 65.50  & 65.57  & 27.06 & 53.60 \\
ConViT Base  & 56.83 & 64.50 & 66.63 & \textbf{27.96} & 53.98 \\
ViT Base 16* & \textbf{75.14} & \textbf{75.14} & \textbf{82.72} & 25.64           & \textbf{64.66} \\ \bottomrule
\end{tabular}
}
\caption{Comparison of different feature extractors without fine-tuning, using a k-NN classifier ($k=5$). The model marked with * is pretrained on Imagenet21K instead of ImageNet1K.}
\label{tab:knn}
\end{table}

\item[Feature Mapping Visualization]
To further enlighten the role of backbones in extracting meaningful and invariant features to deal with DG, we can visualize the distributions in the feature space by projecting them in a two-dimensional space using t-SNE. \cref{fig:tSNE} shows t-SNE visualization for ResNet50 and ConViT Base, pretrained on ImageNet1K and fine-tuned on PACS, targeting the \textit{Art Painting} domain. For each model, we remove the classification head and extract the features for the whole dataset. The more clustered the same class features appear in the t-SNE, the more separable from other classes they are in the original space.

\begin{table}[ht]
\centering
\resizebox{350pt}{!}{%
\begin{tabular}{@{}ccccccc@{}}
\toprule
\textbf{Backbone} & \textbf{Algorithm} & \textbf{PACS} & \textbf{VLCS} & \textbf{Office-Home} & \textbf{Terra Incognita} & \textbf{Overall} \\ \midrule
\multirow{8}{*}{ResNet50 \cite{gulrajani2020search}} & ERM & 85.50 ± 0.20 & 77.50 ± 0.40 & 66.50 ± 0.30 & 46.10 ± 1.80 & 68.90 ± 0.68 \\
 & RSC & 85.20 ± 0.90 & 77.10 ± 0.50 & 65.50 ± 0.90 & 46.60 ± 1.00 & 68.60 ± 0.83 \\
 & Mixup & 84.60 ± 0.60 & 77.40 ± 0.60 & 68.10 ± 0.30 & \textbf{47.90 ± 0.80} & 69.50 ± 0.58 \\
 & CORAL & \textbf{86.20 ± 0.30} & \textbf{78.80 ± 0.60} & \textbf{68.70 ± 0.30} & 47.60 ± 1.00 & \textbf{70.33 ± 0.55} \\
 & MMD & 84.60 ± 0.50 & 77.50 ± 0.90 & 66.30 ± 0.10 & 42.20 ± 1.60 & 67.65 ± 0.78 \\
 & CausIRL CORAL & 85.80 ± 0.10 & 77.50 ± 0.60 & 68.60 ± 0.30 & 47.30 ± 0.80 & 69.80 ± 0.45 \\
 & CausIRL MMD & 84.00 ± 0.80 & 77.60 ± 0.40 & 65.70 ± 0.60 & 46.30 ± 0.90 & 68.40 ± 0.68 \\
 & CAD & 85.20 ± 0.90 & 78.00 ± 0.50 & 67.40 ± 0.20 & 47.30 ± 2.20 & 69.48 ± 0.95 \\ 
 & ADDG \cite{meng2022attention} & 89.2 & - & 72.5 & - & - \\ \midrule
\multirow{8}{*}{DeiT Base 16} & ERM & \textbf{88.10 ± 0.48} & \textbf{79.80 ± 0.32} & {76.35 ± 0.36} & {47.22 ± 0.75} & \textbf{72.87 ± 0.48} \\
 & RSC & 85.37 ± 1.30 & 77.27 ± 0.51 & 76.47 ± 0.28 & 45.41 ± 1.50 & 70.97 ± 0.90 \\
 & Mixup & 85.67 ± 0.61 & 78.25 ± 0.60 & 75.96 ± 0.11 & 46.63 ± 0.49 & 71.32 ± 0.48 \\
 & CORAL & 85.13 ± 0.82 & 78.34 ± 0.86 & 76.48 ± 0.14 & 46.33 ± 1.83 & 71.38 ± 0.93 \\
 & MMD & 87.22 ± 0.28 & 78.71 ± 0.22 & \textbf{77.03 ± 0.10} & \textbf{49.35 ± 1.42} & \textbf{73.08 ± 0.50} \\
 & CausIRL CORAL & 83.86 ± 0.75 & 77.80 ± 0.40 & 76.12 ± 0.04 & 46.73 ± 0.81 & 71.13 ± 0.50 \\
 & CausIRL MMD & 85.46 ± 0.68 & 77.27 ± 0.42 & 76.53 ± 0.42 & 45.77 ± 1.66 & 71.26 ± 0.79 \\
 & CAD & \textbf{87.74 ± 0.62} & 79.28 ± 0.36 & 76.61 ± 0.15 & 47.46 ± 0.64 & \textbf{72.77 ± 0.44} \\ 
 &  ADDG & 75.30 ± 0.34 & 78.28 ± 0.77 & \textbf{77.58 ± 0.30} & 29.14 ± 2.24 & 65.07 ± 0.91 \\ \midrule
\multirow{8}{*}{ConViT Base} & ERM & \textbf{87.27 ± 0.40} & \textbf{80.31 ± 0.67} & 76.51 ± 0.25 & \textbf{46.37 ± 0.89} & \textbf{72.62 ± 0.55} \\
 & RSC & 85.73 ± 0.81 & 79.05 ± 0.61 & 76.77 ± 0.26 & 44.94 ± 1.47 & 71.62 ± 0.79 \\
 & Mixup & 86.00 ± 0.45 & 80.00 ± 0.76 & 76.48 ± 0.16 & 43.95 ± 0.18 & 71.61 ± 0.39 \\
 & CORAL & 86.24 ± 0.24 & 79.62 ± 0.38 & 75.33 ± 0.22 & 44.41 ± 1.33 & 71.40 ± 0.54 \\
 & MMD & 86.84 ± 0.63 & \textbf{80.72 ± 0.55} & \textbf{77.94 ± 0.31} & \textbf{46.78 ± 1.22} & \textbf{73.07 ± 0.68} \\
 & CausIRL CORAL & 84.71 ± 0.31 & 79.14 ± 0.69 & 77.05 ± 0.16 & 45.63 ± 2.03 & 71.63 ± 0.80 \\
 & CausIRL MMD & 86.59 ± 0.96 & \textbf{80.30 ± 0.56} & \textbf{77.92 ± 0.35} & \textbf{46.85 ± 0.59} & \textbf{72.92 ± 0.61} \\
 & CAD & \textbf{87.42 ± 0.66} & 79.99 ± 0.41 & \textbf{77.71 ± 0.09} & \textbf{46.77 ± 3.31} & \textbf{72.97 ± 1.12} \\ 
 &  ADDG & 86.34 ± 0.76 & 79.79 ± 0.30 & 76.29 ± 0.33 & 43.97 ± 1.75 & 71.60 ± 0.78 \\ \midrule
\multirow{8}{*}{ViT Base 16*} & ERM & \textbf{88.48 ± 1.22} & 80.05 ± 0.15 & 81.47 ± 0.21 & 49.77 ± 1.28 & \textbf{74.94 ± 0.72} \\
 & RSC & 86.58 ± 2.14 & 79.59 ± 0.63 & 78.74 ± 0.64 & 40.79 ± 1.41 & 71.42 ± 1.20 \\
 & Mixup & \textbf{88.62 ± 0.54} & \textbf{80.77 ± 1.28} & \textbf{82.93 ± 0.07} & 48.59 ± 0.92 & \textbf{75.23 ± 0.70} \\
 & CORAL & 84.60 ± 1.31 & \textbf{80.89 ± 0.49} & 80.92 ± 0.25 & \textbf{50.58 ± 0.26} & 74.25 ± 0.58 \\
 & MMD & 87.99 ± 0.08 & 79.54 ± 0.37 & 81.71 ± 0.28 & 49.40 ± 2.45 & \textbf{74.66 ± 0.79} \\
 & CausIRL CORAL & \textbf{88.26 ± 1.09} & 80.10 ± 0.91 & 81.73 ± 0.13 & 47.29 ± 2.64 & 74.35 ± 1.19 \\
 & CausIRL MMD & 86.57 ± 1.13 & 79.48 ± 1.12 & 81.62 ± 0.22 & 49.52 ± 0.58 & 74.30 ± 0.76 \\
 & CAD & 87.44 ± 0.53 & 78.79 ± 2.43 & 79.80 ± 0.36 & 39.45 ± 4.15 & 71.37 ± 1.87 \\ 
 &  ADDG & 75.33 ± 0.54 & 77.77 ± 0.32 & 77.72 ± 0.09 & 25.60 ± 0.64 & 64.11 ± 0.40 \\ \bottomrule
\end{tabular}%
}
\caption{Comparison between ERM and three promising DG algorithms on the best-performing backbones of our benchmark. We report the average accuracy over three runs and the associated standard deviation for each model. We highlight in bold the best result for each dataset, including ERM when its accuracy is in the same confidence interval. We include the results achieved by \textsc{DomainBed} with ResNet50 for reference. The model marked with * is pretrained on Imagenet21K instead of ImageNet1K. In \cref{appsec:additional-benchmark}, we report in detail the results obtained for all the domains.}
\label{tab:algorithms}
\end{table}

\cref{fig:resnet50_imagenet} shows how ResNet50 pretrained on ImageNet tends to map together same-domain samples and not same-class ones, being therefore unsuitable for DG without fine-tuning. After the fine-tuning process (\cref{fig:resnet50_art_painting}), the model achieves a better separation of source domain classes. However, many target domain samples are still mapped in the same space, far from the same-class source clusters (e.g. the \textit{Art Painting} guitar example). Similarly to ResNet50, without fine-tuning, domains dominate the features space distribution of ConViT (\cref{fig:convit_base_imagenet}), causing several clusters of the same class, but different domains, to emerge in different locations (e.g. horse samples). However, some same-class samples of more similar domains, such as the guitars of  \textit{Cartoon} and \textit{Art Painting}, are effectively clustered together. The fine-tuning process (\cref{fig:convit_base_art_painting}) distinctly pushes together same-class clusters, resulting in good generalization over the target domain. This analysis suggests that the ConViT backbone is more suited for DG than ResNet50 since it tends to give more similar representations to same-class samples from different domains. Additional feature mapping visualizations are presented in \cref{appsec:additional-model-introspection}.

\item[Self-attention Visualization]
In literature, DG algorithms are often presented with a qualitative analysis, highlighting the regions the network focuses on using interpretation methods such as GradCAM \cite{selvaraju2017grad}. Indeed, heat maps are brought as evidence of their capability to push attention toward more localized and domain-invariant features. Nevertheless, this section shows that competitive backbones with naive ERM can perfectly localize class-discriminative regions. In particular, \cref{fig:attention_visualization} shows the attention maps extracted using the {\fontfamily{qcr}\selectfont [CLS]} token as a query for the different heads in the last layer of the DeiT Base architecture. We provide four random examples for different target domains of PACS showing the same attention head map before and after DG fine-tuning. It is remarkable how naive ERM is able to redirect attention towards more invariant features. Additional attention visualizations are reported in \cref{appsec:additional-model-introspection}.
\end{description}

\subsection{Domain Generalization Algorithms}
\label{sec:algorithms_evaluation}
Domain generalization research mainly focuses on studying non-trivial algorithms to reduce the effect of domain shifts on classification accuracy. However, these algorithms are uniquely proposed in combination with outdated backbones such as ResNet50, ResNet18, or even AlexNet. According to the results in  \cref{tab:baselines}, recent backbones can provide significant improvements compared to ResNet50 with simple ERM. At this point, it is worth determining whether the application of DG algorithms brings a further boost in generalization to our baselines. To do so, we combine some of the most promising and recent algorithms available on \textsc{DomainBed} with three of our best baselines. We evaluate the methods introduced at the beginning of this Section (MMD, CORAL, Mixup, RSC, CAD, CausIRL CORAL, CausIRL MMD, and ADDG) using ViT Base 16, DeiT Base 16, and ConViT Base as backbones and repeating each training three times. \cref{tab:algorithms} reports the obtained results, composed of average accuracy and associated standard deviation. Results obtained with ResNet50 are also reported directly from \textsc{DomainBed} for the same group of datasets as a reference. The only exception is the most recent ADDG, which the authors have not tested on VLCS and Terra Incognita and does not report standard errors.

As highlighted by the values in bold, the overall performance of ERM is equal to or better than other DG algorithms for all the considered datasets and backbones. Indeed, even where another methodology slightly outperforms ERM, the accuracy results mostly fall in the same confidence interval and hence differ very little statistically. We can then conclude from our experimentation that DG algorithms improve generalization properties marginally or even negatively for transformer-based backbones. This outcome extends the recent findings of \textsc{DomainBed} to other baselines and strongly reinforces the belief that choosing an effective backbone is the first step towards filling domain gaps. Adopting an outdated or poorly trained baseline is not the correct way to demonstrate the improvement derived from a DG algorithm. In the next section, we briefly ask ourselves what the reason behind this result is. Moreover, in \cref{appsec:additional-benchmark}, we detail the results obtained for all the domains.

\section{Additional Considerations}
\label{sec:additional_consideration}
\subsection{Are Transformer-based Backbones Better at Generalizing?}
Reflecting on experimental evidence and visual introspection from previous sections, we discuss whether transformer-based backbones are more robust to domain shifts in this paragraph. Undoubtedly, all baseline comparisons of \cref{sec:baseline_benchmark} and features visualizations shown in \cref{fig:tSNE} would suggest a positive answer to this interesting question. In all results and visual representations, self-attention-based models tend to generalize better to unseen domains. Nevertheless, exercising caution and critically analyzing all the variables involved in the process is important. Indeed, such a conclusion only holds leveraging our backbone definition as a function of architecture $\A$, training procedure $\T_{B}$, and data $\D$ (as presented in \cref{sec:dg_theory}). Architecture and training procedure are difficult to disentangle, and there is no guarantee that a training procedure optimal for a specific architecture remains the best for another. Therefore, that implies it is impossible to compare two different architectures directly.
Nevertheless, some recent experimentation on residual architectures with current state-of-the-art training procedures has shed some light on the contribution of $\A$ to the generalization process. Indeed, in \cite{wightman2021resnet}, a vanilla ResNet50 is trained with the approach developed by \cite{touvron2021training}, reaching  80.4\% top-1 accuracy on ImageNet without extra data or distillation. However, ResNet50 A1 performs only slightly better than the original model on our \textsc{Back-to-Bones} testbed, even if there is a difference of over 4\% in ImageNet accuracy. That deviates slightly from the linear correlation described in \cref{sec:baseline_benchmark} and suggests that a transformer-based architecture brings a significant generalization contribution. As further evidence of this trend, ConViT Small has a comparable number of parameters with \cite{wightman2021resnet} and a similar training procedure but outperforms it by more than 4\% on some datasets. Nonetheless, further experimentation can yield more comprehensive results on this interesting aspect of vision transformers.

\begin{table}[t]
\centering
\resizebox{0.4\columnwidth}{!}{%
\begin{tabular}{@{}ccc@{}}
\toprule
\textbf{Baseline} & \textbf{Average} & \textbf{Std Deviation} \\ \midrule
TRM\cite{xu2021learning} & 77.13 & 1.53 \\
MMLD\cite{matsuura2020domain} & 78.70 & - \\
JiGen\cite{carlucci2019domain} & 79.05 & - \\
Epi-FCR\cite{Li_2019_ICCV} & 79.05 & - \\
MASF\cite{dou2019domain} & 79.23 & 0.15 \\
SagNet\cite{nam2021reducing} & 79.26 & - \\
DDAIG\cite{zhou2020deep} & 79.53 & 0.48 \\
D-SAM\cite{d2018domain} & 79.55 & - \\
PAdaIN\cite{nuriel2021permuted} & 79.72 & - \\
MetaReg\cite{balaji2018metareg} & 79.93 & 0.28 \\
RSC\cite{huang2020self} & 79.94 & - \\
\textbf{Ours} & \textbf{80.51} & \textbf{0.29} \\ \bottomrule
\end{tabular}
}
\caption{Comparison between our ResNet18 baseline on PACS and those presented by popular DG works. Our accuracy result outperforms all previous ones, which rarely include statistical information from multiple training iterations. Moreover, these works seldom discuss hyperparameter search procedures and model selection strategies.}
\label{tab:barbuto}
\end{table}

\subsection{On Baseline Selection in Previous Works}
As already stated in \cref{sec:intro}, in the last decade, a plethora of algorithms for domain generalization (DG) has been proposed in the literature, trying to tackle the problem with a wide variety of sophisticated methodologies. Nevertheless, our experimentation highlights that baselines often lack proper optimization. \cref{tab:barbuto} compares our accuracy result with those obtained by several recent works. We evaluate ResNet18 on PACS as this is the most common setup, and our baseline outperforms all those reported in the latest research. Moreover, statistical information is often absent in past DG works, overlooking proper hyperparameter search and model selection strategy discussion. In accordance with the outcomes of \textsc{DomainBed}, we hope to encourage the adoption of rigorous testing procedures, in conjunction with a standard model selection strategy, for transparent research results. With this study, we suggest that new DG algorithms should be analyzed based on adopting well-trained backbones. As a matter of fact, an advantage brought to underpowered baselines can be considered meaningless.

\section{Conclusion and Future Work}
\label{sec:conclusion}
In this paper, we deeply investigate the role of backbones in domain generalization, bringing back to light the fundamental contribution neglected by the community that a competitive feature extractor provides for generalizing to out-of-distribution data. According to our suggested backbone definition, novel architectural solutions such as DeiT, ConViT, and LeViT show remarkable improvements in reducing domain gaps with their intrinsic feature mapping mechanisms and achieve state-of-the-art results in DG with naive ERM and data augmentation only. Hence, we point out that a complete domain generalization study should consider the choice of the backbone as the first step. Moreover, we claim that the advantage of adopting generalization algorithms should be proved using recent and effectively trained feature extractors. 

The enhancement of architectures represents the main road to guide future research on DG. For this reason, we encourage the research community to evaluate novel backbones on the proposed testbed \textsc{Back-to-Bones} to keep updated a benchmark dedicated to the DG problem. With this work, we do not add a methodology to the long list but ponder the current situation surrounding DG works, trying to shift towards more effective research.
From a broader perspective, our research points out the fundamental role of backbones in DG. However, we did not thoroughly examine the specific backbone components responsible for the observed correlation between source and target accuracy. Indeed, architecture, backbone training procedure, and data could contribute differently to the measured generalization capabilities. Therefore, we believe that besides collecting additional results with the proposed benchmark, further studies aim at evaluating the role of backbone components in DG.


\section*{Acknowledgments}
This work has been developed with the contribution of the Politecnico di Torino Interdepartmental Centre for Service Robotics (PIC4SeR)\footnote{\url{https://pic4ser.polito.it}} and SmartData@Polito\footnote{\url{https://smartdata.polito.it}}.

\bibliographystyle{abbrv}  
\bibliography{biblio}  

\begin{thebibliography}{10}

\bibitem{albuquerque2020improving}
I.~Albuquerque, N.~Naik, J.~Li, N.~Keskar, and R.~Socher.
\newblock Improving out-of-distribution generalization via multi-task
  self-supervised pretraining.
\newblock {\em arXiv preprint arXiv:2003.13525}, 2020.

\bibitem{arjovsky2019invariant}
M.~Arjovsky, L.~Bottou, I.~Gulrajani, and D.~Lopez-Paz.
\newblock Invariant risk minimization.
\newblock {\em arXiv preprint arXiv:1907.02893}, 2019.

\bibitem{balaji2018metareg}
Y.~Balaji, S.~Sankaranarayanan, and R.~Chellappa.
\newblock Metareg: Towards domain generalization using meta-regularization.
\newblock {\em Advances in Neural Information Processing Systems},
  31:998--1008, 2018.

\bibitem{Beery_2018_ECCV}
S.~Beery, G.~Van~Horn, and P.~Perona.
\newblock Recognition in terra incognita.
\newblock In {\em Proceedings of the European Conference on Computer Vision
  (ECCV)}, 2018.

\bibitem{blanchard2011generalizing}
G.~Blanchard, G.~Lee, and C.~Scott.
\newblock Generalizing from several related classification tasks to a new
  unlabeled sample.
\newblock {\em Advances in neural information processing systems},
  24:2178--2186, 2011.

\bibitem{bucci2022self}
S.~Bucci, A.~D’Innocente, Y.~Liao, F.~M. Carlucci, B.~Caputo, and T.~Tommasi.
\newblock Self-supervised learning across domains.
\newblock {\em IEEE Transactions on Pattern Analysis and Machine Intelligence},
  44(9):5516--5528, 2022.

\bibitem{carlucci2019domain}
F.~M. Carlucci, A.~D'Innocente, S.~Bucci, B.~Caputo, and T.~Tommasi.
\newblock Domain generalization by solving jigsaw puzzles.
\newblock In {\em Proceedings of the IEEE/CVF Conference on Computer Vision and
  Pattern Recognition}, pages 2229--2238, 2019.

\bibitem{chen2023domain}
S.~Chen, L.~Wang, Z.~Hong, and X.~Yang.
\newblock Domain generalization by joint-product distribution alignment.
\newblock {\em Pattern Recognition}, 134:109086, 2023.

\bibitem{chevalley2022invariant}
M.~Chevalley, C.~Bunne, A.~Krause, and S.~Bauer.
\newblock Invariant causal mechanisms through distribution matching.
\newblock {\em arXiv preprint arXiv:2206.11646}, 2022.

\bibitem{csurka2017domain}
G.~Csurka.
\newblock {\em Domain adaptation in computer vision applications}.
\newblock Springer, 2017.

\bibitem{dai2018dark}
D.~Dai and L.~Van~Gool.
\newblock Dark model adaptation: Semantic image segmentation from daytime to
  nighttime.
\newblock In {\em 2018 21st International Conference on Intelligent
  Transportation Systems (ITSC)}, pages 3819--3824. IEEE, 2018.

\bibitem{d2021convit}
S.~D'Ascoli, H.~Touvron, M.~L. Leavitt, A.~S. Morcos, G.~Biroli, and L.~Sagun.
\newblock Convit: Improving vision transformers with soft convolutional
  inductive biases.
\newblock In M.~Meila and T.~Zhang, editors, {\em Proceedings of the 38th
  International Conference on Machine Learning}, volume 139 of {\em Proceedings
  of Machine Learning Research}, pages 2286--2296. PMLR, 18--24 Jul 2021.

\bibitem{deng2009imagenet}
J.~Deng, W.~Dong, R.~Socher, L.-J. Li, K.~Li, and L.~Fei-Fei.
\newblock Imagenet: A large-scale hierarchical image database.
\newblock In {\em 2009 IEEE conference on computer vision and pattern
  recognition}, pages 248--255. Ieee, 2009.

\bibitem{dosovitskiy2021an}
A.~Dosovitskiy, L.~Beyer, A.~Kolesnikov, D.~Weissenborn, X.~Zhai,
  T.~Unterthiner, M.~Dehghani, M.~Minderer, G.~Heigold, S.~Gelly, J.~Uszkoreit,
  and N.~Houlsby.
\newblock An image is worth 16x16 words: Transformers for image recognition at
  scale.
\newblock In {\em International Conference on Learning Representations}, 2021.

\bibitem{dou2019domain}
Q.~Dou, D.~Coelho~de Castro, K.~Kamnitsas, and B.~Glocker.
\newblock Domain generalization via model-agnostic learning of semantic
  features.
\newblock {\em Advances in Neural Information Processing Systems},
  32:6450--6461, 2019.

\bibitem{d2018domain}
A.~D’Innocente and B.~Caputo.
\newblock Domain generalization with domain-specific aggregation modules.
\newblock In {\em German Conference on Pattern Recognition}, pages 187--198.
  Springer, 2018.

\bibitem{elharrouss2022backbones}
O.~Elharrouss, Y.~Akbari, N.~Almaadeed, and S.~Al-Maadeed.
\newblock Backbones-review: Feature extraction networks for deep learning and
  deep reinforcement learning approaches.
\newblock {\em arXiv preprint arXiv:2206.08016}, 2022.

\bibitem{Fang_2013_ICCV}
C.~Fang, Y.~Xu, and D.~N. Rockmore.
\newblock Unbiased metric learning: On the utilization of multiple datasets and
  web images for softening bias.
\newblock In {\em Proceedings of the IEEE International Conference on Computer
  Vision (ICCV)}, 2013.

\bibitem{ganin2016domain}
Y.~Ganin, E.~Ustinova, H.~Ajakan, P.~Germain, H.~Larochelle, F.~Laviolette,
  M.~Marchand, and V.~Lempitsky.
\newblock Domain-adversarial training of neural networks.
\newblock {\em The journal of machine learning research}, 17(1):2096--2030,
  2016.

\bibitem{gatys2015texture}
L.~Gatys, A.~S. Ecker, and M.~Bethge.
\newblock Texture synthesis using convolutional neural networks.
\newblock In C.~Cortes, N.~Lawrence, D.~Lee, M.~Sugiyama, and R.~Garnett,
  editors, {\em Advances in Neural Information Processing Systems}, volume~28.
  Curran Associates, Inc., 2015.

\bibitem{geirhos2018imagenet}
R.~Geirhos, P.~Rubisch, C.~Michaelis, M.~Bethge, F.~A. Wichmann, and
  W.~Brendel.
\newblock Imagenet-trained {CNN}s are biased towards texture; increasing shape
  bias improves accuracy and robustness.
\newblock In {\em International Conference on Learning Representations}, 2019.

\bibitem{ghifary2015domain}
M.~Ghifary, W.~B. Kleijn, M.~Zhang, and D.~Balduzzi.
\newblock Domain generalization for object recognition with multi-task
  autoencoders.
\newblock In {\em Proceedings of the IEEE international conference on computer
  vision}, pages 2551--2559, 2015.

\bibitem{graham2021levit}
B.~Graham, A.~El-Nouby, H.~Touvron, P.~Stock, A.~Joulin, H.~J{\'e}gou, and
  M.~Douze.
\newblock Levit: a vision transformer in convnet's clothing for faster
  inference.
\newblock In {\em Proceedings of the IEEE/CVF international conference on
  computer vision}, pages 12259--12269, 2021.

\bibitem{gulrajani2020search}
I.~Gulrajani and D.~Lopez-Paz.
\newblock In search of lost domain generalization.
\newblock In {\em International Conference on Learning Representations}.
  Computer Vision Foundation, 2021.

\bibitem{He_2016_CVPR}
K.~He, X.~Zhang, S.~Ren, and J.~Sun.
\newblock Deep residual learning for image recognition.
\newblock In {\em Proceedings of the IEEE Conference on Computer Vision and
  Pattern Recognition (CVPR)}, 2016.

\bibitem{he2016deep}
K.~He, X.~Zhang, S.~Ren, and J.~Sun.
\newblock Deep residual learning for image recognition.
\newblock In {\em Proceedings of the IEEE conference on computer vision and
  pattern recognition}, pages 770--778, 2016.

\bibitem{howard2017mobilenets}
A.~G. Howard, M.~Zhu, B.~Chen, D.~Kalenichenko, W.~Wang, T.~Weyand,
  M.~Andreetto, and H.~Adam.
\newblock Mobilenets: Efficient convolutional neural networks for mobile vision
  applications.
\newblock {\em arXiv preprint arXiv:1704.04861}, 2017.

\bibitem{huang2017densely}
G.~Huang, Z.~Liu, L.~Van Der~Maaten, and K.~Q. Weinberger.
\newblock Densely connected convolutional networks.
\newblock In {\em Proceedings of the IEEE conference on computer vision and
  pattern recognition}, pages 4700--4708, 2017.

\bibitem{huang2020self}
Z.~Huang, H.~Wang, E.~P. Xing, and D.~Huang.
\newblock Self-challenging improves cross-domain generalization.
\newblock In {\em Computer Vision--ECCV 2020: 16th European Conference,
  Glasgow, UK, August 23--28, 2020, Proceedings, Part II 16}, pages 124--140.
  Springer, 2020.

\bibitem{kim2021selfreg}
D.~Kim, Y.~Yoo, S.~Park, J.~Kim, and J.~Lee.
\newblock Selfreg: Self-supervised contrastive regularization for domain
  generalization.
\newblock In {\em Proceedings of the IEEE/CVF International Conference on
  Computer Vision}, pages 9619--9628, 2021.

\bibitem{krizhevsky2012imagenet}
A.~Krizhevsky, I.~Sutskever, and G.~E. Hinton.
\newblock Imagenet classification with deep convolutional neural networks.
\newblock {\em Advances in neural information processing systems},
  25:1097--1105, 2012.

\bibitem{Li_2017_ICCV}
D.~Li, Y.~Yang, Y.-Z. Song, and T.~M. Hospedales.
\newblock Deeper, broader and artier domain generalization.
\newblock In {\em Proceedings of the IEEE International Conference on Computer
  Vision (ICCV)}, 2017.

\bibitem{li2018learning}
D.~Li, Y.~Yang, Y.-Z. Song, and T.~M. Hospedales.
\newblock Learning to generalize: Meta-learning for domain generalization.
\newblock In {\em Thirty-Second AAAI Conference on Artificial Intelligence},
  2018.

\bibitem{Li_2019_ICCV}
D.~Li, J.~Zhang, Y.~Yang, C.~Liu, Y.-Z. Song, and T.~M. Hospedales.
\newblock Episodic training for domain generalization.
\newblock In {\em Proceedings of the IEEE/CVF International Conference on
  Computer Vision (ICCV)}, 2019.

\bibitem{li2018domain}
H.~Li, S.~J. Pan, S.~Wang, and A.~C. Kot.
\newblock Domain generalization with adversarial feature learning.
\newblock In {\em Proceedings of the IEEE Conference on Computer Vision and
  Pattern Recognition}, pages 5400--5409, 2018.

\bibitem{loquercio2019deep}
A.~Loquercio, E.~Kaufmann, R.~Ranftl, A.~Dosovitskiy, V.~Koltun, and
  D.~Scaramuzza.
\newblock Deep drone racing: From simulation to reality with domain
  randomization.
\newblock {\em IEEE Transactions on Robotics}, 36(1):1--14, 2019.

\bibitem{matsuura2020domain}
T.~Matsuura and T.~Harada.
\newblock Domain generalization using a mixture of multiple latent domains.
\newblock In {\em Proceedings of the AAAI Conference on Artificial
  Intelligence}, volume~34, pages 11749--11756, 2020.

\bibitem{meng2022attention}
R.~Meng, X.~Li, W.~Chen, S.~Yang, J.~Song, X.~Wang, L.~Zhang, M.~Song, D.~Xie,
  and S.~Pu.
\newblock Attention diversification for domain generalization.
\newblock In {\em Computer Vision--ECCV 2022: 17th European Conference, Tel
  Aviv, Israel, October 23--27, 2022, Proceedings, Part XXXIV}, pages 322--340.
  Springer, 2022.

\bibitem{motiian2017unified}
S.~Motiian, M.~Piccirilli, D.~A. Adjeroh, and G.~Doretto.
\newblock Unified deep supervised domain adaptation and generalization.
\newblock In {\em Proceedings of the IEEE international conference on computer
  vision}, pages 5715--5725, 2017.

\bibitem{mozifian2020intervention}
M.~Mozifian, A.~Zhang, J.~Pineau, and D.~Meger.
\newblock Intervention design for effective sim2real transfer.
\newblock {\em arXiv preprint arXiv:2012.02055}, 2020.

\bibitem{muandet2013domain}
K.~Muandet, D.~Balduzzi, and B.~Sch{\"o}lkopf.
\newblock Domain generalization via invariant feature representation.
\newblock In {\em International Conference on Machine Learning}, pages 10--18.
  PMLR, 2013.

\bibitem{nam2021reducing}
H.~Nam, H.~Lee, J.~Park, W.~Yoon, and D.~Yoo.
\newblock Reducing domain gap by reducing style bias.
\newblock In {\em Proceedings of the IEEE/CVF Conference on Computer Vision and
  Pattern Recognition}, pages 8690--8699, 2021.

\bibitem{nuriel2021permuted}
O.~Nuriel, S.~Benaim, and L.~Wolf.
\newblock Permuted adain: Reducing the bias towards global statistics in image
  classification.
\newblock In {\em Proceedings of the IEEE/CVF Conference on Computer Vision and
  Pattern Recognition}, pages 9482--9491, 2021.

\bibitem{Peng_2019_ICCV}
X.~Peng, Q.~Bai, X.~Xia, Z.~Huang, K.~Saenko, and B.~Wang.
\newblock Moment matching for multi-source domain adaptation.
\newblock In {\em Proceedings of the IEEE/CVF International Conference on
  Computer Vision (ICCV)}. Computer Vision Foundation, 2019.

\bibitem{rahman2020correlation}
M.~M. Rahman, C.~Fookes, M.~Baktashmotlagh, and S.~Sridharan.
\newblock Correlation-aware adversarial domain adaptation and generalization.
\newblock {\em Pattern Recognition}, 100:107124, 2020.

\bibitem{ruan2022optimal}
Y.~Ruan, Y.~Dubois, and C.~J. Maddison.
\newblock Optimal representations for covariate shift.
\newblock In {\em International Conference on Learning Representations}, 2022.

\bibitem{sagawa2019distributionally}
S.~Sagawa, P.~W. Koh, T.~B. Hashimoto, and P.~Liang.
\newblock Distributionally robust neural networks.
\newblock In {\em International Conference on Learning Representations}, 2020.

\bibitem{segu2023batch}
M.~Segu, A.~Tonioni, and F.~Tombari.
\newblock Batch normalization embeddings for deep domain generalization.
\newblock {\em Pattern Recognition}, 135:109115, 2023.

\bibitem{selvaraju2017grad}
R.~R. Selvaraju, M.~Cogswell, A.~Das, R.~Vedantam, D.~Parikh, and D.~Batra.
\newblock Grad-cam: Visual explanations from deep networks via gradient-based
  localization.
\newblock In {\em Proceedings of the IEEE international conference on computer
  vision}, pages 618--626, 2017.

\bibitem{shahtalebi2021sand}
S.~Shahtalebi, J.-C. Gagnon-Audet, T.~Laleh, M.~Faramarzi, K.~Ahuja, and
  I.~Rish.
\newblock Sand-mask: An enhanced gradient masking strategy for the discovery of
  invariances in domain generalization.
\newblock {\em arXiv preprint arXiv:2106.02266}, 2021.

\bibitem{shankar2018generalizing}
S.~Shankar, V.~Piratla, S.~Chakrabarti, S.~Chaudhuri, P.~Jyothi, and
  S.~Sarawagi.
\newblock Generalizing across domains via cross-gradient training.
\newblock In {\em International Conference on Learning Representations}, 2018.

\bibitem{simonyan2014very}
K.~Simonyan and A.~Zisserman.
\newblock Very deep convolutional networks for large-scale image recognition.
\newblock {\em arXiv preprint arXiv:1409.1556}, 2014.

\bibitem{sun2016deep}
B.~Sun and K.~Saenko.
\newblock Deep {CORAL}: Correlation alignment for deep domain adaptation.
\newblock In {\em European conference on computer vision}, pages 443--450.
  Springer, 2016.

\bibitem{szegedy2015going}
C.~Szegedy, W.~Liu, Y.~Jia, P.~Sermanet, S.~Reed, D.~Anguelov, D.~Erhan,
  V.~Vanhoucke, and A.~Rabinovich.
\newblock Going deeper with convolutions.
\newblock In {\em Proceedings of the IEEE conference on computer vision and
  pattern recognition}, pages 1--9, 2015.

\bibitem{tan2019efficientnet}
M.~Tan and Q.~Le.
\newblock Efficientnet: Rethinking model scaling for convolutional neural
  networks.
\newblock In {\em International Conference on Machine Learning}, pages
  6105--6114. PMLR, 2019.

\bibitem{pmlr-v97-tan19a}
M.~Tan and Q.~Le.
\newblock {E}fficient{N}et: Rethinking model scaling for convolutional neural
  networks.
\newblock In {\em Proceedings of the 36th International Conference on Machine
  Learning}, volume~97 of {\em Proceedings of Machine Learning Research}, pages
  6105--6114. PMLR, 2019.

\bibitem{tobin2018domain}
J.~Tobin, L.~Biewald, R.~Duan, M.~Andrychowicz, A.~Handa, V.~Kumar, B.~McGrew,
  A.~Ray, J.~Schneider, P.~Welinder, et~al.
\newblock Domain randomization and generative models for robotic grasping.
\newblock In {\em 2018 IEEE/RSJ International Conference on Intelligent Robots
  and Systems (IROS)}, pages 3482--3489. IEEE, 2018.

\bibitem{tobin2017domain}
J.~Tobin, R.~Fong, A.~Ray, J.~Schneider, W.~Zaremba, and P.~Abbeel.
\newblock Domain randomization for transferring deep neural networks from
  simulation to the real world.
\newblock In {\em 2017 IEEE/RSJ international conference on intelligent robots
  and systems (IROS)}, pages 23--30. IEEE, 2017.

\bibitem{touvron2021training}
H.~Touvron, M.~Cord, M.~Douze, F.~Massa, A.~Sablayrolles, and H.~J{\'e}gou.
\newblock Training data-efficient image transformers \& distillation through
  attention.
\newblock In {\em Proceedings of the 38th International Conference on Machine
  Learning}, volume 139 of {\em Proceedings of Machine Learning Research},
  pages 10347--10357. PMLR, 2021.

\bibitem{valiant2013probably}
L.~Valiant.
\newblock {\em Probably Approximately Correct: Nature's Algorithms for Learning
  and Prospering in a Complex World}.
\newblock Basic Books (AZ), 2013.

\bibitem{van2008visualizing}
L.~Van~der Maaten and G.~Hinton.
\newblock Visualizing data using {t-SNE}.
\newblock {\em Journal of machine learning research}, 9(11), 2008.

\bibitem{vapnik1999overview}
V.~N. Vapnik.
\newblock An overview of statistical learning theory.
\newblock {\em IEEE transactions on neural networks}, 10(5):988--999, 1999.

\bibitem{vaswani2017attention}
A.~Vaswani, N.~Shazeer, N.~Parmar, J.~Uszkoreit, L.~Jones, A.~N. Gomez,
  {\L}.~Kaiser, and I.~Polosukhin.
\newblock Attention is all you need.
\newblock In {\em Advances in neural information processing systems}, pages
  5998--6008, 2017.

\bibitem{Venkateswara_2017_CVPR}
H.~Venkateswara, J.~Eusebio, S.~Chakraborty, and S.~Panchanathan.
\newblock Deep hashing network for unsupervised domain adaptation.
\newblock In {\em Proceedings of the IEEE Conference on Computer Vision and
  Pattern Recognition (CVPR)}, 2017.

\bibitem{volk2019towards}
G.~Volk, S.~M{\"u}ller, A.~von Bernuth, D.~Hospach, and O.~Bringmann.
\newblock Towards robust cnn-based object detection through augmentation with
  synthetic rain variations.
\newblock In {\em 2019 IEEE Intelligent Transportation Systems Conference
  (ITSC)}, pages 285--292. IEEE, 2019.

\bibitem{volpi2018generalizing}
R.~Volpi, H.~Namkoong, O.~Sener, J.~C. Duchi, V.~Murino, and S.~Savarese.
\newblock Generalizing to unseen domains via adversarial data augmentation.
\newblock In {\em Advances in Neural Information Processing Systems},
  volume~31, 2018.

\bibitem{wightman2021resnet}
R.~Wightman, H.~Touvron, and H.~Jegou.
\newblock Resnet strikes back: An improved training procedure in timm.
\newblock In {\em NeurIPS 2021 Workshop on ImageNet: Past, Present, and
  Future}, 2021.

\bibitem{xu2021learning}
Y.~Xu and T.~Jaakkola.
\newblock Learning representations that support robust transfer of predictors.
\newblock {\em arXiv preprint arxiv:2110.09940}, 2021.

\bibitem{yan2020improve}
S.~Yan, H.~Song, N.~Li, L.~Zou, and L.~Ren.
\newblock Improve unsupervised domain adaptation with mixup training.
\newblock {\em arXiv preprint arXiv:2001.00677}, 2020.

\bibitem{zhang2020adaptive}
M.~M. Zhang, H.~Marklund, N.~Dhawan, A.~Gupta, S.~Levine, and C.~Finn.
\newblock Adaptive risk minimization: A meta-learning approach for tackling
  group shift.
\newblock In {\em International Conference on Learning Representations}, 2020.

\bibitem{zhou2021domain}
K.~Zhou, Z.~Liu, Y.~Qiao, T.~Xiang, and C.~C. Loy.
\newblock Domain generalization: A survey.
\newblock {\em IEEE Transactions on Pattern Analysis and Machine Intelligence},
  45(4):4396--4415, 2023.

\bibitem{zhou2020deep}
K.~Zhou, Y.~Yang, T.~Hospedales, and T.~Xiang.
\newblock Deep domain-adversarial image generation for domain generalisation.
\newblock In {\em Proceedings of the AAAI Conference on Artificial
  Intelligence}, volume~34, pages 13025--13032, 2020.

\end{thebibliography}
\clearpage

\appendix

\section{Additional Benchmark Results}
\label{appsec:additional-benchmark}
\subsection{Baseline Benchmark}
This section includes additional results from our benchmark of different backbones, described in \cref{sec:baseline_benchmark}. In particular, \cref{tab:supp_baselines} reports accuracy results for each target domain, further highlighting the differences between different models. Besides some oscillations given by peculiar domains, the trend of novel and more performing backbones overcoming outdated ones is present for each target split. Each value is recorded as the mean over three independent runs, along with its standard deviation.

\subsection{Domain Generalization Algorithms}
As for the baseline benchmark, we report more detailed results for the experimentation described in \cref{sec:algorithms_evaluation}. In particular, \cref{tab:meth} includes accuracy results for each target domain to highlight the strengths and the weaknesses of the examined algorithms. In general, the evaluated methodologies do not significantly benefit DG accuracy, performing similarly or even worse than naive ERM for nearly all target domains. Moreover, the unreliability of DG algorithms highlighted by our experimentation is further compounded by the fact that target labels are rarely available in a real scenario. That prevents a direct check of the effectiveness of the procedure, leading to the adoption of more robust and trustworthy solutions. Indeed, ERM is easy to implement and consistently achieves remarkable accuracy results. Each value is recorded as the mean over three independent runs, along with its standard deviation.

\begin{table}[hbt]
\resizebox{\columnwidth}{!}{%
\begin{subtable}[h]{0.5\columnwidth}
\centering
\resizebox{\columnwidth}{!}{%
\begin{tabular}{@{}ccccc@{}}
\toprule
\textbf{Backbone} & \textbf{Photo} & \textbf{Art Painting} & \textbf{Cartoon} & \textbf{Sketch} \\ \midrule
ResNet18       & 94.03 ± 0.49 & 79.65 ± 1.74 & 75.71 ± 0.75 & 72.63 ± 1.53 \\
ResNet50       & 94.53 ± 0.54 & 82.86 ± 2.42 & 76.83 ± 4.70 & 81.18 ± 0.68 \\
ResNet50 A1    & 97.84 ± 0.66 & 85.87 ± 0.60 & 74.43 ± 1.33 & 79.94 ± 0.86 \\ \midrule
EfficientNetB0 & 95.43 ± 0.23 & 82.29 ± 1.00 & 80.69 ± 1.55 & 83.45 ± 2.14 \\
EfficientNetB2 & 96.61 ± 0.35 & 85.30 ± 1.44 & 82.75 ± 1.09 & 83.44 ± 3.21 \\
EfficientNetB3 & 95.89 ± 0.60 & 83.82 ± 0.54 & 81.93 ± 1.23 & 85.20 ± 1.18 \\ \midrule
DeiT Small 16  & 98.38 ± 1.25 & 87.58 ± 2.13 & 81.32 ± 1.96 & 77.60 ± 0.88 \\
DeiT Base 16   & 99.38 ± 0.15 & 90.74 ± 0.75 & 82.75 ± 1.07 & 79.52 ± 1.54 \\
ConViT Small   & 99.16 ± 0.16 & 91.23 ± 0.57 & 81.58 ± 1.45 & 76.43 ± 1.38 \\
ConViT Base    & 99.18 ± 0.18 & 91.05 ± 0.75 & 81.39 ± 1.61 & 77.47 ± 1.45 \\
LeViT Base     & 98.04 ± 0.55 & 85.20 ± 2.67 & 86.21 ± 1.59 & 80.76 ± 2.15 \\ \midrule
ViT Small 16*  & 99.40 ± 0.12 & 89.94 ± 0.52 & 80.60 ± 2.46 & 64.40 ± 1.20 \\
ViT Base 32*   & 99.30 ± 0.33 & 89.63 ± 0.28 & 80.22 ± 0.91 & 66.84 ± 4.60 \\
ViT Base 16*   & 99.50 ± 0.17 & 93.25 ± 1.09 & 85.52 ± 2.64 & 75.67 ± 2.29 \\ \bottomrule
\end{tabular}}
\caption{PACS}
\label{tab:supp_baselines_pacs}
\end{subtable}
\hspace{10pt}
\begin{subtable}[h]{0.5\columnwidth}
\centering
\resizebox{\columnwidth}{!}{%
\begin{tabular}{@{}ccccc@{}}
\toprule
\textbf{Backbone} & \textbf{Caltech} & \textbf{Labelme} & \textbf{Pascal} & \textbf{Sun} \\ \midrule
ResNet18 & 95.60 ± 0.18 & 62.55 ± 1.29 & 72.80 ± 1.90 & 67.60 ± 1.63 \\
ResNet50 & 96.09 ± 1.36 & 64.47 ± 1.72 & 73.43 ± 3.34 & 70.83 ± 3.25 \\
ResNet50 A1       & 98.89 ± 0.16     & 63.23 ± 0.77     & 77.64 ± 1.84    & 73.72 ± 0.55 \\ \midrule
EfficientNetB0    & 96.84 ± 0.99     & 61.76 ± 0.40     & 70.43 ± 1.16    & 71.62 ± 1.01 \\
EfficientNetB2    & 97.97 ± 1.27     & 63.94 ± 0.95     & 71.11 ± 1.52    & 68.73 ± 0.62 \\
EfficientNetB3    & 97.36 ± 0.41     & 63.68 ± 0.99     & 76.32 ± 1.51    & 75.20 ± 2.09 \\ \midrule
DeiT Small 16     & 97.53 ± 0.38     & 64.91 ± 0.52     & 79.58 ± 1.21    & 75.85 ± 1.02 \\ 
DeiT Base 16      & 97.79 ± 0.18     & 65.24 ± 0.29     & 78.06 ± 2.17    & 78.11 ± 1.41 \\
ConViT Small      & 97.95 ± 0.31     & 64.98 ± 0.51     & 80.33 ± 1.07    & 76.72 ± 0.67 \\
ConViT Base       & 97.95 ± 0.37     & 65.78 ± 0.36     & 79.22 ± 2.63    & 78.28 ± 2.01 \\
LeViT Base        & 98.19 ± 0.30     & 64.52 ± 1.13     & 76.57 ± 0.74    & 76.38 ± 1.30 \\ \midrule
ViT Small 16*     & 97.90 ± 0.41     & 64.86 ± 1.99     & 79.41 ± 2.72    & 77.67 ± 0.96 \\
ViT Base 32*      & 98.78 ± 0.71     & 64.66 ± 0.24     & 75.97 ± 1.40    & 74.46 ± 2.79 \\
ViT Base 16*      & 97.69 ± 0.59     & 65.42 ± 2.28     & 79.55 ± 2.64    & 77.53 ± 0.41 \\ \bottomrule
\end{tabular}}
\caption{VLCS}
\label{tab:supp_baselines_vlcs}
\end{subtable}
}
\resizebox{\columnwidth}{!}{%
\begin{subtable}[h]{0.5\columnwidth}
\centering
\vspace{20pt}
\resizebox{\columnwidth}{!}{%
\begin{tabular}{@{}ccccc@{}}
\toprule
\textbf{Backbone} & \textbf{Product} & \textbf{Art} & \textbf{Clipart} & \textbf{Real World} \\ \midrule
ResNet18       & 73.88 ± 0.36 & 55.42 ± 0.91 & 52.73 ± 0.39 & 73.46 ± 0.20 \\
ResNet50       & 77.16 ± 0.28 & 63.23 ± 0.46 & 56.26 ± 0.20 & 78.51 ± 1.00 \\
ResNet50 A1    & 79.88 ± 0.57 & 69.28 ± 0.44 & 57.63 ± 0.68 & 83.11 ± 0.32 \\ \midrule
EfficientNetB0 & 75.96 ± 0.28 & 60.91 ± 0.72 & 54.56 ± 1.39 & 77.65 ± 0.06 \\
EfficientNetB2 & 78.52 ± 0.67 & 62.99 ± 0.66 & 56.27 ± 0.45 & 79.62 ± 0.51 \\
EfficientNetB3 & 79.37 ± 0.43 & 64.50 ± 0.83 & 55.13 ± 1.04 & 80.38 ± 0.41 \\ \midrule
DeiT Small 16  & 79.91 ± 0.55 & 69.30 ± 0.93 & 56.74 ± 0.63 & 82.16 ± 0.21 \\
DeiT Base 16   & 83.55 ± 0.37 & 75.22 ± 0.55 & 61.07 ± 0.52 & 85.55 ± 0.64 \\
ConViT Small   & 80.69 ± 0.37 & 72.45 ± 0.83 & 58.69 ± 0.17 & 83.79 ± 0.44 \\
ConViT Base    & 83.21 ± 0.58 & 74.49 ± 0.50 & 62.58 ± 1.44 & 85.77 ± 0.31 \\
LeViT Base     & 83.23 ± 0.14 & 72.85 ± 0.91 & 60.55 ± 0.56 & 84.02 ± 0.66 \\ \midrule
ViT Small 16*   & 84.79 ± 0.55 & 76.32 ± 0.82 & 60.50 ± 0.72 & 87.38 ± 0.58 \\
ViT Base 32*    & 83.92 ± 0.37 & 75.28 ± 0.33 & 60.95 ± 0.71 & 87.21 ± 0.15 \\
ViT Base 16*    & 88.39 ± 0.35 & 79.93 ± 0.87 & 67.71 ± 0.36 & 89.85 ± 0.89 \\ \bottomrule
\end{tabular}}
\caption{Office-Home}
\label{tab:supp_baselines_officehome}
\end{subtable}
\hspace{10pt}
\begin{subtable}[h]{0.5\columnwidth}
\centering
\vspace{20pt}
\resizebox{\columnwidth}{!}{%
\begin{tabular}{@{}ccccc@{}}
\toprule
\textbf{Backbone} & \textbf{L100} & \textbf{L38} & \textbf{L43} & \textbf{L46} \\ \midrule
ResNet18          & 44.48 ± 2.71  & 35.95 ± 2.87 & 49.26 ± 2.08 & 34.03 ± 1.50 \\
ResNet50          & 53.19 ± 5.66  & 41.57 ± 2.48 & 54.10 ± 1.33 & 40.43 ± 1.42 \\
ResNet50 A1       & 48.22 ± 1.90  & 39.41 ± 2.77 & 45.50 ± 2.49 & 35.77 ± 0.84 \\ \midrule
EfficientNetB0    & 44.73 ± 3.15  & 41.40 ± 4.13 & 54.14 ± 0.83 & 38.76 ± 0.99 \\
EfficientNetB2    & 41.92 ± 0.75  & 41.42 ± 4.02 & 55.06 ± 1.74 & 36.80 ± 3.08 \\
EfficientNetB3    & 48.93 ± 2.76  & 37.92 ± 3.69 & 58.27 ± 1.08 & 37.67 ± 2.40 \\ \midrule
DeiT Small 16     & 53.11 ± 2.52  & 30.64 ± 5.40 & 50.79 ± 2.44 & 39.07 ± 1.76 \\
DeiT Base 16      & 58.53 ± 1.07  & 35.93 ± 1.42 & 52.57 ± 2.82 & 41.83 ± 2.12 \\
ConViT Small      & 53.55 ± 1.06  & 36.41 ± 0.44 & 53.93 ± 2.09 & 39.43 ± 0.65 \\
ConViT Base       & 52.17 ± 4.05  & 32.54 ± 3.26 & 57.30 ± 0.27 & 43.50 ± 2.44 \\
LeViT Base        & 55.02 ± 3.48  & 36.23 ± 2.34 & 55.37 ± 0.92 & 36.11 ± 1.52 \\ \midrule
ViT Small 16*      & 54.10 ± 4.17  & 33.70 ± 1.04 & 50.15 ± 2.23 & 38.52 ± 2.25 \\
ViT Base 32*       & 33.33 ± 3.86  & 28.84 ± 2.64 & 52.57 ± 2.82 & 32.09 ± 0.94 \\
ViT Base 16*       & 58.65 ± 4.18  & 41.14 ± 2.12 & 56.83 ± 1.39 & 42.47 ± 2.62 \\ \bottomrule
\end{tabular}}
\caption{Terra Incognita}
\label{tab:supp_baselines_terraincognita}
\end{subtable}
}
\caption{Baselines comparison of different backbones on the four considered DG datasets. We report the average accuracy over three runs and the associated standard deviation for each model. The models marked with * are pretrained on Imagenet21K instead of ImageNet1K.}
\label{tab:supp_baselines}
\end{table}

\begin{table}[ht]
\resizebox{\columnwidth}{!}{%
\begin{subtable}[h]{0.5\columnwidth}
\centering
\resizebox{\columnwidth}{!}{%
\begin{tabular}{@{}cccccc@{}}
\toprule
\textbf{Backbone} & \textbf{Algorithm} & \textbf{Photo} & \textbf{Art Painting} & \textbf{Cartoon} & \textbf{Sketch} \\ \midrule
\multirow{8}{*}{DeiT Base 16} & ERM & 99.38 ± 0.15 & 90.74 ± 0.75 & 82.75 ± 1.07 & 79.52 ± 1.54 \\
 & RSC & 99.20 ± 0.12 & 87.65 ± 0.51 & 78.46 ± 3.18 & 76.19 ± 1.77 \\
 & Mixup & 99.28 ± 0.16 & 87.52 ± 0.81 & 77.20 ± 1.16 & 78.66 ± 2.36 \\
 & CORAL & 99.44 ± 0.15 & 87.97 ± 0.32 & 75.74 ± 3.67 & 77.38 ± 0.92 \\
 & MMD & 99.48 ± 0.03 & 89.44 ± 0.85 & 79.75 ± 0.65 & 80.20 ± 1.15 \\
 & CausIRL CORAL & 99.28 ± 0.26 & 86.56 ± 0.79 & 72.94 ± 1.35 & 76.65 ± 1.82 \\
 & CausIRL MMD & 99.04 ± 0.36 & 87.75 ± 1.02 & 77.69 ± 1.11 & 77.36 ± 2.70 \\
 & CAD & 99.54 ± 0.19 & 90.10 ± 0.44 & 81.08 ± 1.58 & 80.23 ± 1.42 \\ 
 &  ADDG & 96.87 ± 0.88  & 87.50 ± 1.18  & 72.65 ± 2.22  & 44.16 ± 2.15\\ \midrule
\multirow{8}{*}{ConViT Base} & ERM & 99.18 ± 0.18 & 91.05 ± 0.75 & 81.39 ± 1.61 & 77.47 ± 1.45 \\
 & RSC & 98.36 ± 1.33 & 89.96 ± 0.93 & 78.37 ± 2.05 & 76.22 ± 2.06 \\
 & Mixup & 99.32 ± 0.30 & 90.79 ± 0.71 & 78.94 ± 1.05 & 74.94 ± 0.55 \\
 & CORAL & 99.50 ± 0.09 & 90.67 ± 0.64 & 78.06 ± 0.64 & 76.73 ± 2.12 \\
 & MMD & 99.54 ± 0.12 & 90.84 ± 0.62 & 80.19 ± 1.10 & 76.77 ± 1.82 \\
 & CausIRL CORAL & 99.18 ± 0.30 & 88.92 ± 1.14 & 75.41 ± 0.60 & 75.32 ± 1.70 \\
 & CausIRL MMD & 99.50 ± 0.09 & 90.67 ± 1.15 & 79.45 ± 0.33 & 76.74 ± 2.43 \\
 & CAD & 99.48 ± 0.17 & 90.63 ± 0.83 & 82.76 ± 0.58 & 76.80 ± 1.36 \\ 
 &  ADDG & 99.34 ± 0.12  & 90.14 ± 1.16  & 80.79 ± 0.86  & 75.11 ± 1.92 \\ \midrule
\multirow{8}{*}{ViT Base 16*} & ERM & 99.50 ± 0.17 & 93.25 ± 1.09 & 85.52 ± 2.64 & 75.67 ± 2.29 \\
 & RSC & 98.94 ± 0.40 & 92.07 ± 2.34 & 86.03 ± 1.18 & 69.26 ± 7.69 \\
 & Mixup & 99.60 ± 0.12 & 95.04 ± 0.32 & 87.17 ± 0.87 & 72.67 ± 0.94 \\
 & CORAL & 98.38 ± 1.00 & 93.08 ± 0.73 & 81.20 ± 1.39 & 65.75 ± 4.21 \\
 & MMD & 99.36 ± 0.18 & 94.56 ± 0.41 & 84.88 ± 2.18 & 73.14 ± 1.58 \\
 & CausIRL CORAL & 99.50 ± 0.12 & 94.07 ± 0.42 & 86.36 ± 0.14 & 73.11 ± 4.03 \\
 & CausIRL MMD & 99.42 ± 0.09 & 93.57 ± 1.27 & 83.96 ± 2.10 & 69.34 ± 2.53 \\
 & CAD & 99.52 ± 0.06 & 93.85 ± 0.91 & 84.91 ± 0.45 & 71.47 ± 1.65 \\ 
 &  ADDG & 97.80 ± 0.54  & 88.35 ± 0.57  & 72.22 ± 0.62  & 42.96 ± 1.58\\ \bottomrule
\end{tabular}
}
\caption{PACS}
\label{tab:meth_pacs}
\end{subtable}
\hspace{10pt}
\begin{subtable}[h]{0.5\columnwidth}
\centering
\resizebox{\columnwidth}{!}{%
\begin{tabular}{@{}cccccc@{}}
\toprule
\textbf{Backbone} & \textbf{Algorithm} & \textbf{Caltech} & \textbf{Labelme} & \textbf{Pascal} & \textbf{Sun} \\ \midrule
\multirow{8}{*}{DeiT Base 16} & ERM & 97.79 ± 0.18 & 65.24 ± 0.29 & 78.06 ± 2.17 & 78.11 ± 1.41 \\
 & RSC & 97.95 ± 0.96 & 64.95 ± 0.26 & 71.42 ± 0.58 & 74.77 ± 0.79 \\
 & Mixup & 98.42 ± 0.30 & 63.72 ± 0.60 & 74.59 ± 2.19 & 76.25 ± 0.52 \\
 & CORAL & 97.74 ± 0.55 & 65.32 ± 0.95 & 73.55 ± 1.80 & 76.74 ± 0.43 \\
 & MMD & 98.09 ± 0.88 & 64.29 ± 1.13 & 74.74 ± 0.90 & 77.71 ± 0.08 \\
 & CausIRL CORAL & 98.87 ± 0.35 & 62.64 ± 0.33 & 73.95 ± 1.31 & 75.72 ± 0.96 \\
 & CausIRL MMD & 97.36 ± 0.59 & 63.67 ± 2.00 & 71.97 ± 0.48 & 76.09 ± 0.31 \\
 & CAD & 96.89 ± 0.77 & 65.30 ± 0.91 & 77.80 ± 1.56 & 77.14 ± 1.25 \\ 
 &  ADDG & 99.22 ± 0.12  & 65.19 ± 0.89  & 77.17 ± 1.06  & 71.54 ± 1.99 \\ \midrule
\multirow{8}{*}{ConViT Base} & ERM & 97.95 ± 0.37 & 65.78 ± 0.36 & 79.22 ± 2.63 & 78.28 ± 2.01 \\
 & RSC & 98.19 ± 0.29 & 65.50 ± 1.21 & 77.89 ± 0.86 & 74.63 ± 1.54 \\
 & Mixup & 99.22 ± 0.14 & 64.57 ± 0.85 & 79.17 ± 2.81 & 77.04 ± 0.38 \\
 & CORAL & 98.35 ± 0.25 & 66.49 ± 0.75 & 77.14 ± 1.06 & 76.51 ± 0.41 \\
 & MMD & 97.92 ± 0.55 & 68.03 ± 0.23 & 79.44 ± 1.54 & 77.47 ± 0.17 \\
 & CausIRL CORAL & 99.20 ± 0.29 & 64.59 ± 1.07 & 77.53 ± 2.78 & 75.25 ± 0.31 \\
 & CausIRL MMD & 98.61 ± 0.34 & 65.86 ± 0.58 & 79.55 ± 1.40 & 77.16 ± 0.68 \\
 & CAD & 97.67 ± 0.12 & 65.49 ± 0.58 & 80.01 ± 1.44 & 76.80 ± 0.61 \\ 
 &  ADDG & 98.45 ± 0.92  & 65.28 ± 1.00  & 79.02 ± 1.21  & 76.43 ± 1.38 \\ \midrule
\multirow{8}{*}{ViT Base 16*} & ERM & 97.69 ± 0.59 & 65.42 ± 2.28 & 79.55 ± 2.64 & 77.53 ± 0.41 \\
 & RSC & 98.23 ± 1.84 & 65.21 ± 0.81 & 78.02 ± 1.64 & 76.89 ± 1.38 \\
 & Mixup & 97.46 ± 0.46 & 66.87 ± 0.31 & 80.26 ± 4.70 & 78.51 ± 0.64 \\
 & CORAL & 98.78 ± 0.46 & 67.24 ± 0.47 & 78.33 ± 0.38 & 79.20 ± 1.20 \\
 & MMD & 98.07 ± 0.55 & 64.92 ± 0.64 & 77.98 ± 2.01 & 77.20 ± 0.15 \\
 & CausIRL CORAL & 97.03 ± 1.10 & 65.89 ± 0.73 & 79.93 ± 2.10 & 77.54 ± 3.33 \\
 & CausIRL MMD & 97.67 ± 0.40 & 65.59 ± 0.59 & 79.16 ± 3.52 & 75.50 ± 1.19 \\
 & CAD & 98.09 ± 0.12 & 60.30 ± 8.14 & 79.38 ± 2.61 & 77.37 ± 1.25 \\ 
 &  ADDG & 99.43 ± 0.08  & 63.30 ± 2.83  & 76.41 ± 1.20  & 71.93 ± 1.89 \\ \bottomrule 
\end{tabular}
}
\caption{VLCS}
\label{tab:meth_vlcs}
\end{subtable}
}
\resizebox{\columnwidth}{!}{%
\begin{subtable}[h]{0.5\columnwidth}
\centering
\vspace{20pt}
\resizebox{\columnwidth}{!}{%
\begin{tabular}{@{}cccccc@{}}
\toprule
\textbf{Backbone} & \textbf{Algorithm} & \textbf{Art} & \textbf{Clipart} & \textbf{Product} & \textbf{Real World} \\ \midrule
\multirow{8}{*}{DeiT Base 16} & ERM & 75.22 ± 0.55 & 61.07 ± 0.52 & 83.55 ± 0.37 & 85.55 ± 0.64 \\
 & RSC & 75.11 ± 0.11 & 61.70 ± 1.26 & 83.07 ± 0.40 & 83.41 ± 0.22 \\
 & Mixup & 75.10 ± 0.75 & 60.83 ± 0.93 & 81.62 ± 0.20 & 81.37 ± 0.25 \\
 & CORAL & 74.60 ± 0.77 & 61.53 ± 0.06 & 83.31 ± 0.47 & 83.45 ± 0.28 \\
 & MMD & 75.53 ± 0.18 & 62.44 ± 0.91 & 83.36 ± 0.31 & 86.78 ± 0.14 \\
 & CausIRL CORAL & 74.69 ± 0.22 & 61.02 ± 0.25 & 82.56 ± 0.07 & 86.22 ± 0.22 \\
 & CausIRL MMD & 74.50 ± 0.73 & 61.79 ± 0.91 & 83.43 ± 0.15 & 86.40 ± 0.25 \\
 & CAD & 75.20 ± 0.40 & 61.76 ± 0.94 & 83.52 ± 0.15 & 85.95 ± 0.05 \\ 
 &  ADDG & 76.62 ± 0.80  & 60.62 ± 1.58  & 85.82 ± 0.38  & 87.25 ± 0.27 \\ \midrule
\multirow{8}{*}{ConViT Base} & ERM & 74.49 ± 0.50 & 62.58 ± 1.44 & 83.21 ± 0.58 & 85.77 ± 0.31 \\
 & RSC & 75.95 ± 0.17 & 63.97 ± 0.32 & 83.49 ± 0.64 & 83.68 ± 0.36 \\
 & Mixup & 75.79 ± 0.60 & 62.95 ± 0.11 & 81.67 ± 0.13 & 85.52 ± 0.18 \\
 & CORAL & 75.71 ± 0.54 & 63.37 ± 0.24 & 81.18 ± 0.19 & 81.05 ± 0.52 \\
 & MMD & 76.46 ± 0.68 & 64.71 ± 0.64 & 83.79 ± 0.56 & 86.80 ± 0.24 \\
 & CausIRL CORAL & 75.16 ± 0.54 & 63.54 ± 0.73 & 82.86 ± 0.12 & 86.20 ± 0.24 \\
 & CausIRL MMD & 76.47 ± 0.25 & 64.74 ± 0.68 & 83.63 ± 0.28 & 86.84 ± 0.34 \\
 & CAD & 76.43 ± 0.37 & 64.31 ± 0.25 & 83.83 ± 0.08 & 86.26 ± 0.35 \\ 
 &  ADDG & 74.42 ± 0.63  & 62.86 ± 0.54  & 82.41 ± 0.44  & 85.46 ± 0.30 \\ \midrule
\multirow{8}{*}{ViT Base 16*} & ERM & 79.93 ± 0.87 & 67.71 ± 0.36 & 88.39 ± 0.35 & 89.85 ± 0.89 \\
 & RSC & 76.99 ± 0.54 & 66.32 ± 0.34 & 85.42 ± 2.33 & 86.21 ± 0.53 \\
 & Mixup & 82.05 ± 0.31 & 69.70 ± 0.28 & 89.17 ± 0.34 & 90.77 ± 0.34 \\
 & CORAL & 79.95 ± 0.34 & 67.14 ± 0.44 & 87.98 ± 0.56 & 89.63 ± 0.23 \\
 & MMD & 80.48 ± 0.28 & 68.33 ± 0.59 & 88.12 ± 0.53 & 89.89 ± 0.26 \\
 & CausIRL CORAL & 80.16 ± 0.56 & 68.54 ± 1.48 & 88.30 ± 0.35 & 89.94 ± 0.29 \\
 & CausIRL MMD & 80.44 ± 0.17 & 68.22 ± 0.65 & 88.19 ± 0.10 & 89.65 ± 0.32 \\
 & CAD & 78.13 ± 0.60 & 66.89 ± 0.64 & 86.19 ± 0.24 & 87.99 ± 0.57 \\ 
 &  ADDG & 76.93 ± 0.33  & 60.44 ± 0.34  & 85.87 ± 0.45  & 87.62 ± 0.30\\ \bottomrule
\end{tabular}
}
\caption{Office-Home}
\label{tab:meth_officehome}
\end{subtable}
\hspace{10pt}
\begin{subtable}[h]{0.5\columnwidth}
\centering
\vspace{20pt}
\resizebox{\columnwidth}{!}{%
\begin{tabular}{@{}cccccc@{}}
\toprule
\textbf{Backbone} & \textbf{Algorithm} & \textbf{L100} & \textbf{L38} & \textbf{L43} & \textbf{L46} \\ \midrule
\multirow{8}{*}{DeiT Base 16} & ERM & 58.53 ± 1.07 & 35.93 ± 1.42 & 52.57 ± 2.82 & 41.83 ± 2.12 \\
 & RSC & 56.55 ± 2.31 & 29.77 ± 5.20 & 51.91 ± 2.01 & 43.42 ± 0.73 \\
 & Mixup & 48.40 ± 1.09 & 35.62 ± 0.43 & 54.73 ± 1.02 & 47.76 ± 1.46 \\
 & CORAL & 52.28 ± 3.75 & 35.06 ± 3.15 & 52.32 ± 1.04 & 45.65 ± 2.33 \\
 & MMD & 57.11 ± 4.33 & 38.27 ± 2.38 & 56.38 ± 2.71 & 45.63 ± 2.58 \\
 & CausIRL CORAL & 48.73 ± 3.87 & 37.55 ± 2.86 & 54.17 ± 1.15 & 46.46 ± 1.32 \\
 & CausIRL MMD & 53.38 ± 3.61 & 33.51 ± 3.77 & 51.95 ± 1.46 & 44.23 ± 0.97 \\
 & CAD & 53.94 ± 1.92 & 38.07 ± 1.90 & 53.69 ± 1.79 & 44.11 ± 0.86 \\ 
 &  ADDG & 23.78 ± 2.17  & 39.85 ± 7.85  & 29.76 ± 3.00  & 23.18 ± 3.28 \\ \midrule
\multirow{8}{*}{ConViT Base} & ERM & 52.17 ± 4.05 & 32.54 ± 3.26 & 57.30 ± 0.27 & 43.50 ± 2.44 \\
 & RSC & 48.27 ± 4.20 & 31.87 ± 1.45 & 55.82 ± 0.72 & 43.80 ± 1.03 \\
 & Mixup & 44.45 ± 1.81 & 29.82 ± 0.88 & 55.82 ± 0.28 & 45.73 ± 2.19 \\
 & CORAL & 45.91 ± 5.60 & 31.17 ± 4.07 & 56.03 ± 2.38 & 44.52 ± 1.37 \\
 & MMD & 48.39 ± 0.30 & 33.57 ± 3.10 & 58.02 ± 1.15 & 47.13 ± 2.89 \\
 & CausIRL CORAL & 43.63 ± 1.91 & 35.67 ± 6.21 & 57.33 ± 1.03 & 45.88 ± 0.57 \\
 & CausIRL MMD & 47.04 ± 3.78 & 36.39 ± 0.84 & 58.44 ± 0.65 & 45.54 ± 1.22 \\
 & CAD & 50.88 ± 4.26 & 34.43 ± 6.64 & 57.78 ± 1.78 & 44.01 ± 1.44 \\ 
 &  ADDG & 43.57 ± 0.67  & 30.47 ± 4.60  & 56.32 ± 0.94  & 45.51 ± 2.66 \\ \midrule
\multirow{8}{*}{ViT Base 16*} & ERM & 58.65 ± 4.18 & 41.14 ± 2.12 & 56.83 ± 1.39 & 42.47 ± 2.62 \\
 & RSC & 48.83 ± 5.00 & 30.84 ± 6.25 & 52.39 ± 1.01 & 31.10 ± 3.84 \\
 & Mixup & 58.50 ± 2.43 & 38.26 ± 1.16 & 57.12 ± 2.32 & 40.49 ± 0.92 \\
 & CORAL & 59.77 ± 2.14 & 44.58 ± 0.92 & 56.31 ± 4.21 & 41.65 ± 1.70 \\
 & MMD & 60.50 ± 2.22 & 42.88 ± 4.99 & 54.90 ± 1.69 & 39.31 ± 5.26 \\
 & CausIRL CORAL & 55.03 ± 8.22 & 39.82 ± 2.94 & 53.98 ± 1.72 & 40.33 ± 1.73 \\
 & CausIRL MMD & 56.71 ± 2.52 & 42.57 ± 1.45 & 55.53 ± 1.40 & 43.28 ± 2.85 \\
 & CAD & 47.53 ± 11.7 & 32.31 ± 1.90 & 46.65 ± 6.57 & 31.29 ± 0.85 \\ 
 &  ADDG & 23.91 ± 1.88 & 29.30 ± 2.84  & 26.73 ± 3.86  & 22.48 ± 3.86\\ \bottomrule
\end{tabular}
}
\caption{Terra Incognita}
\label{tab:meth_terraincognita}
\end{subtable}
}
\caption{Comparison of different DG algorithms on the three best backbones from our benchmark, covering each domain of the four considered datasets. We report the average accuracy over three runs and the associated standard deviation for each model. The model marked with * is pretrained on Imagenet21K instead of ImageNet1K.}
\label{tab:meth}
\end{table}

\section{Additional Model Introspection}
\label{appsec:additional-model-introspection}
\subsection{ResNet50 vs ResNet50 A1}
\begin{figure}[ht]
    \begin{subfigure}{0.8\linewidth}
        \centering
        \includegraphics[width=0.6\columnwidth, trim={10px 10px 10px 10px}, clip]
        {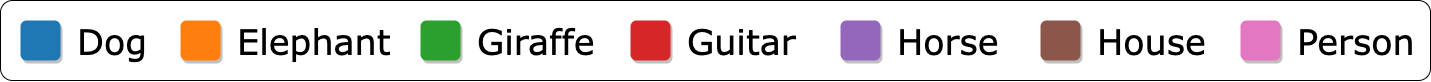}
        \label{fig:supp_legend}
    \end{subfigure}
    \centering
  \resizebox{0.7\columnwidth}{!}{%
  \begin{subfigure}{0.5\linewidth}
    \centering
    \includegraphics[width=\columnwidth, trim={40px 45px 30px 40px}, clip]
                    {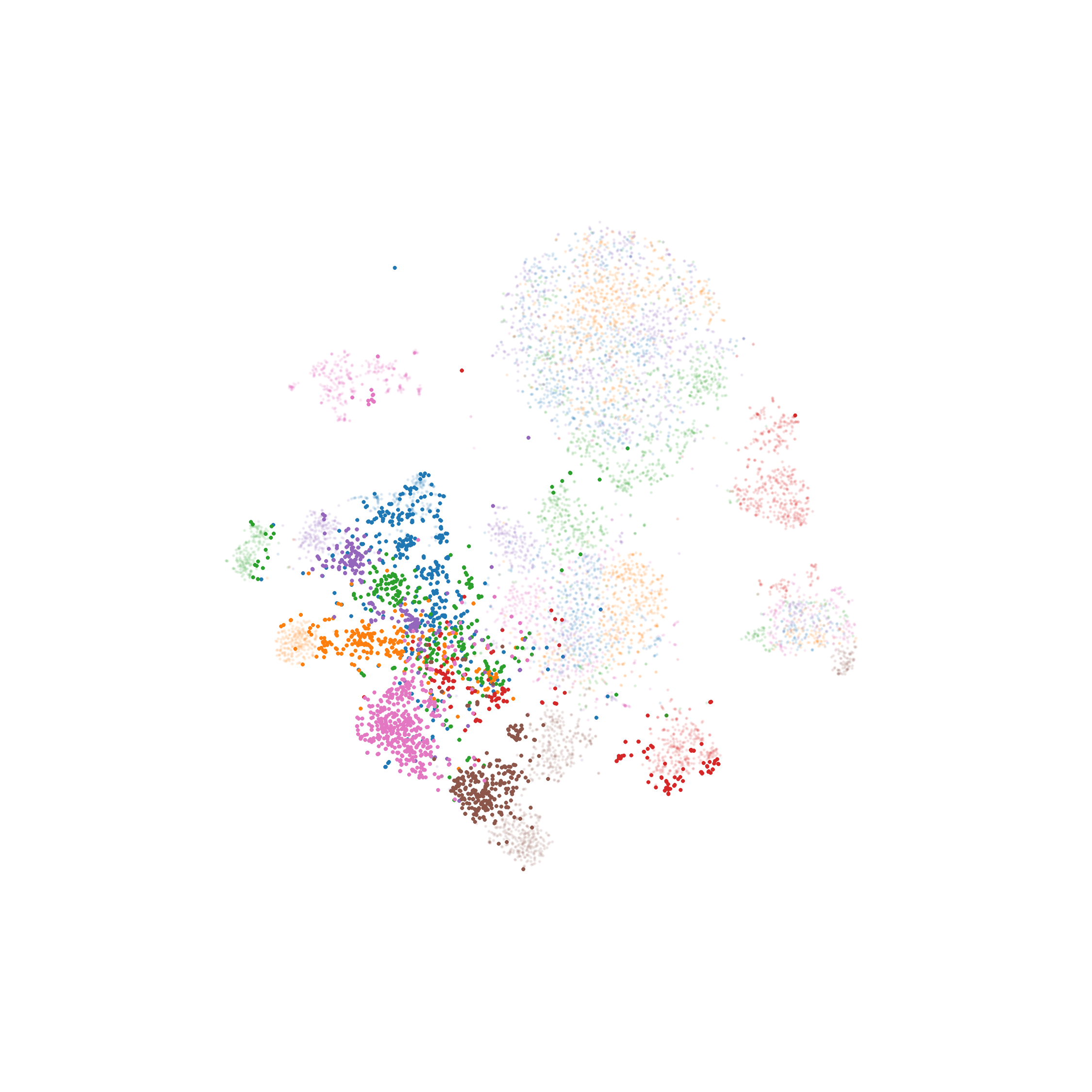}
    \caption{ResNet50 (ImageNet1K)}
    \label{fig:supp_resnet50_imagenet}
  \end{subfigure}
  \begin{subfigure}{0.5\linewidth}
    \centering
    \includegraphics[width=\columnwidth, trim={40 45px 30px 40px}, clip]
                    {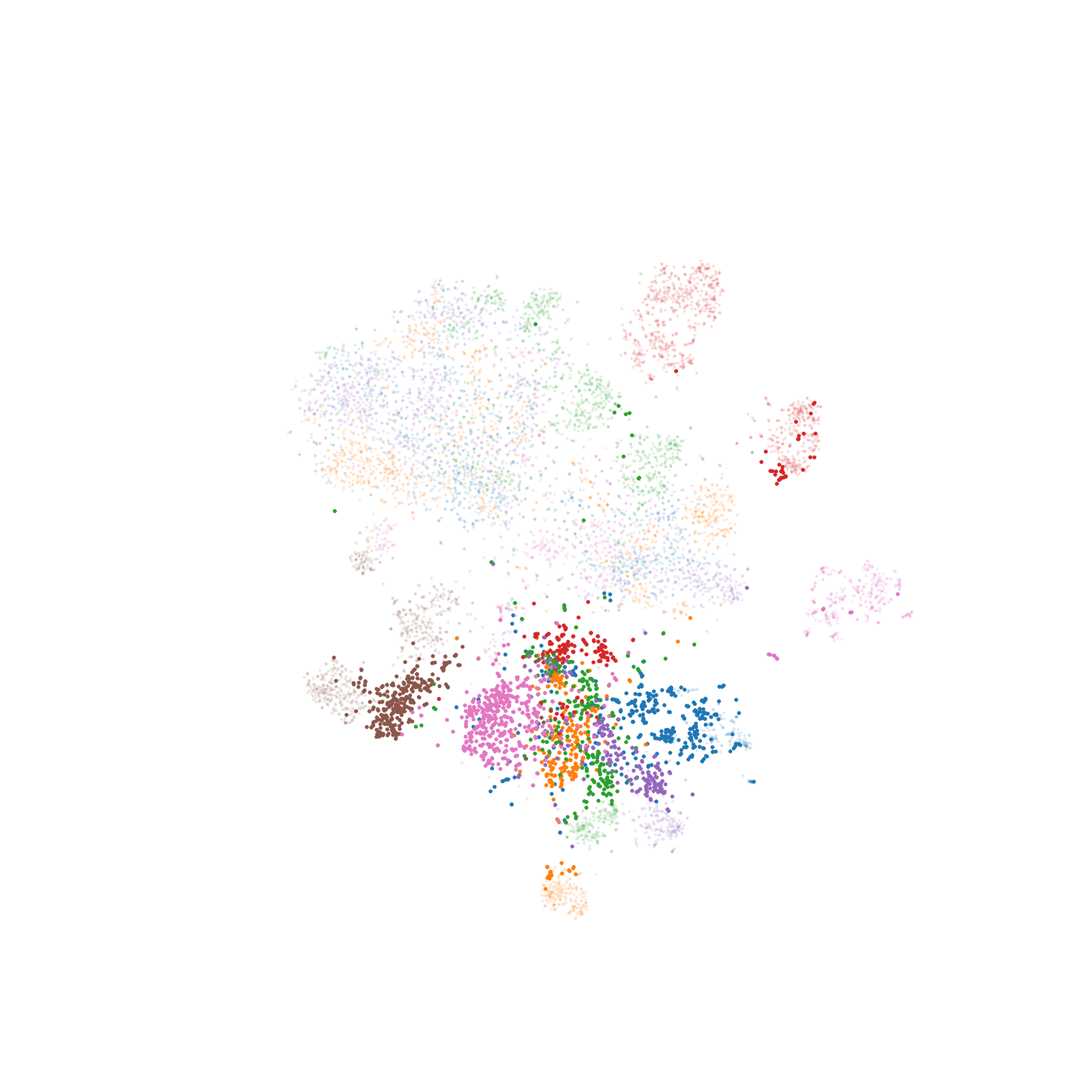}
    \caption{ResNet50 A1 (ImageNet1K)}
    \label{fig:resnet50_rsb}
  \end{subfigure}
  }
  \resizebox{0.7\columnwidth}{!}{%
  \begin{subfigure}{0.5\linewidth}
    \centering
    \includegraphics[width=\columnwidth, trim={40 45px 30px 40px}, clip]
                    {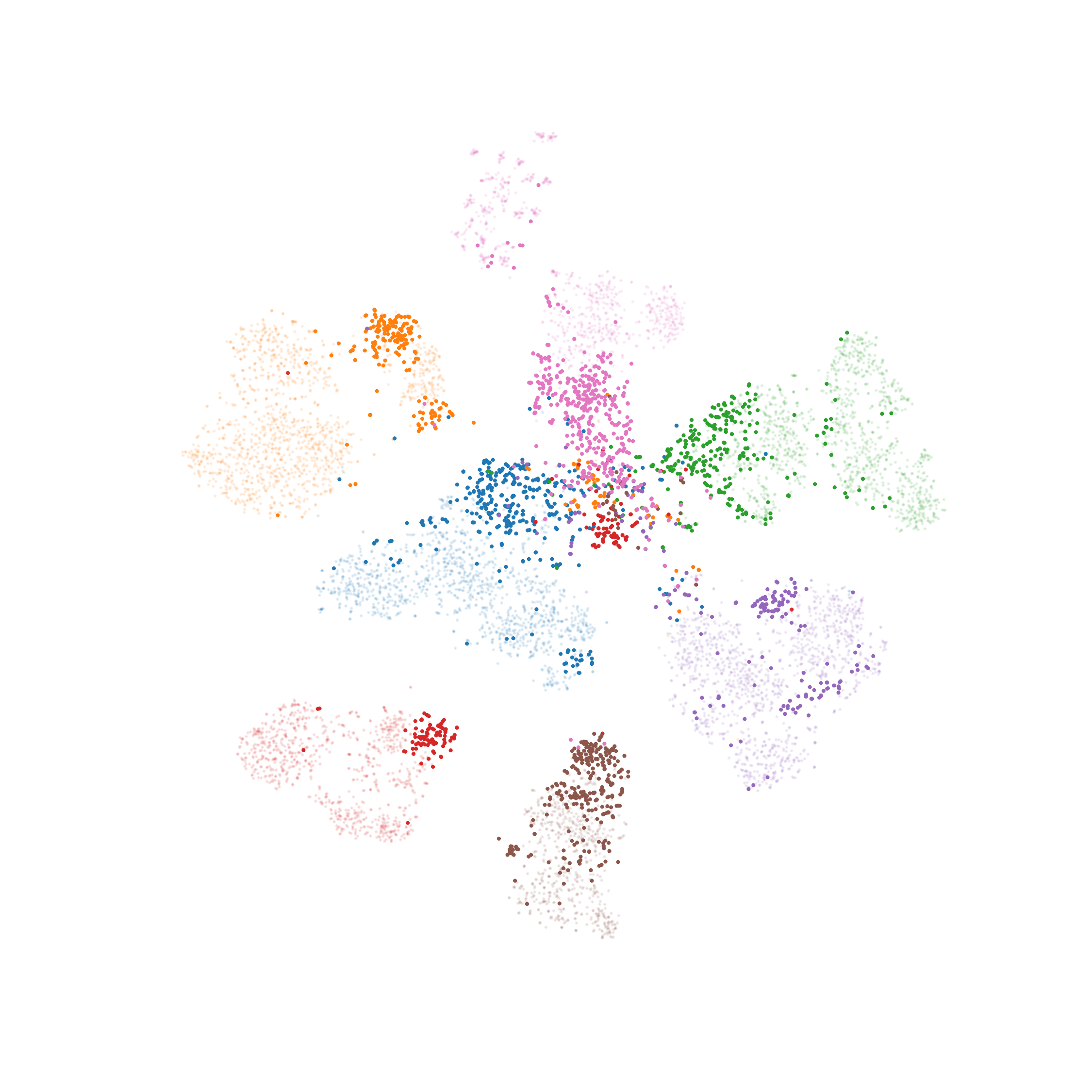}
    \caption{ResNet50 ($\mathcal{S}$: [\textit{P}, \textit{C}, \textit{S}]
                          $\rightarrow$ $T$: \textit{A})}
    \label{fig:supp_resnet50_art_painting}
  \end{subfigure}
  \begin{subfigure}{0.5\linewidth}
    \centering
    \includegraphics[width=\columnwidth, trim={40 45px 30px 40px}, clip]
                    {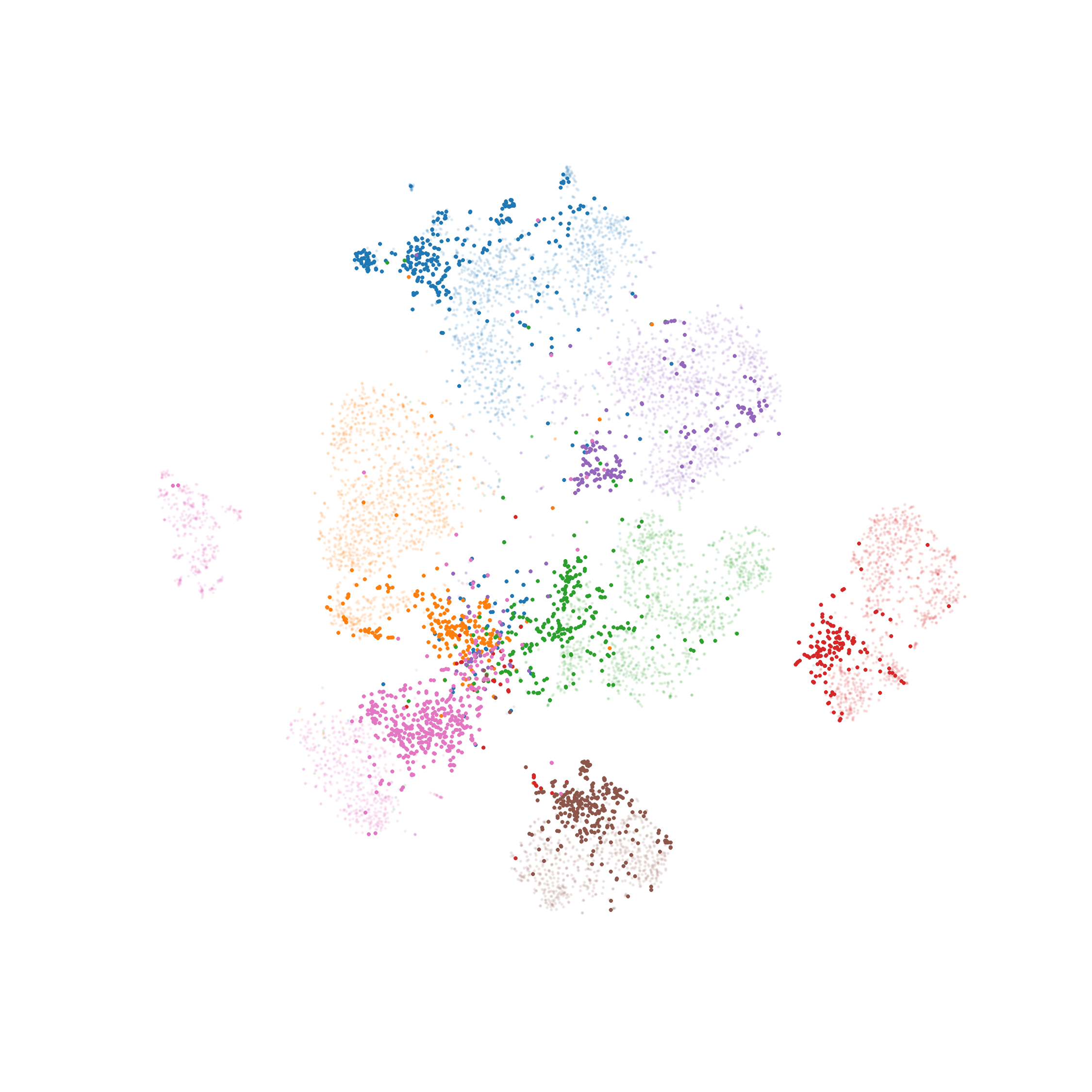}
    \caption{ResNet50 A1 ($\mathcal{S}$: [\textit{P}, \textit{C}, \textit{S}]
                          $\rightarrow$ $T$: \textit{A})}
    \label{fig:resnet50_rsb_art_painting}
  \end{subfigure}
  }
  \caption{PACS features extracted with ResNet50 and ResNet50 A1 projected on 2D space with t-SNE. Target domain \textit{Art Painting} samples are highlighted. Even though ResNet50 A1 has a higher starting accuracy on ImageNet1K, the two backbones have comparable feature space distributions.}
  \label{fig:resnet50_v_resnet50a1}
\end{figure}

ResNet50 A1 \cite{wightman2021resnet} is a retrained version of the popular ResNet50, exploiting the most recent techniques in data augmentation and hyperparameter search, leading to an increased top-1 accuracy on ImageNet1K test set of 80.4\%. \cref{fig:resnet50_v_resnet50a1} compares the t-SNE visualization of the two models. Even though ResNet50 A1 starts with a remarkable advantage in terms of ImageNet accuracy, the two backbones generate comparable feature distributions. In particular, both models tend to separate samples by domain and not by class without fine-tuning (see \cref{fig:supp_resnet50_imagenet}), which does not favor DG. After retraining on three source domains (\textit{Photo}, \textit{Cartoon} and \textit{Sketch}), same-class clusters emerge, still with a certain overlapping over the \textit{Art Painting} target domain.

\subsection{Feature Mapping Visualization}
\begin{figure}[ht]
  \centering
  \begin{subfigure}{\linewidth}
      \centering
      \includegraphics[width=\linewidth, trim={0px 0px 0px 0px}]
      {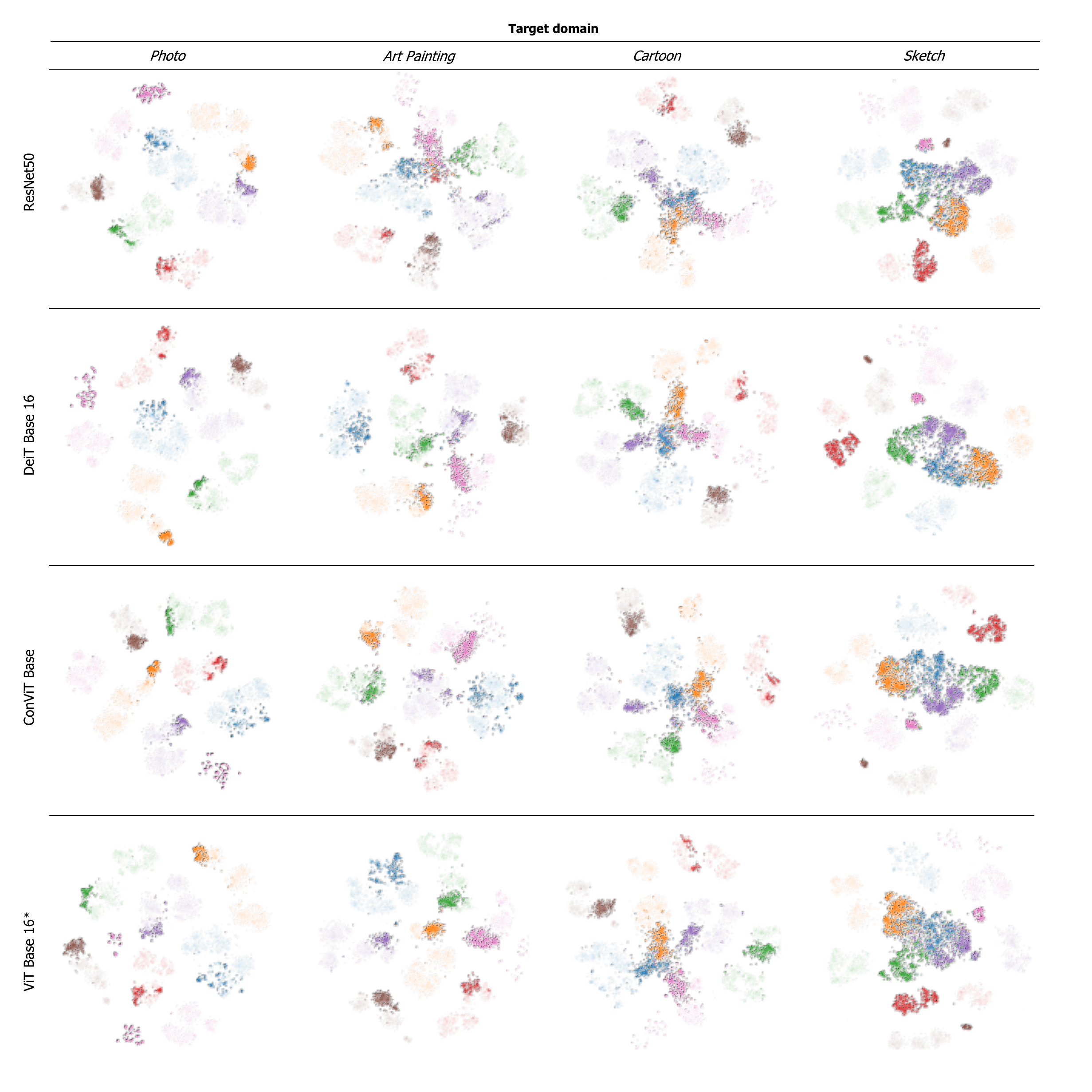}
  \end{subfigure}
  \begin{subfigure}{0.8\linewidth}
        \centering
        \includegraphics[width=0.7\columnwidth, trim={10px 10px 10px 10px}, clip]
        {media/legend_drawio_hor.png}
    \end{subfigure}
  \caption{The t-SNE representation of features extracted from all PACS target domains, with ResNet50 and several transformer-based networks, shows how better the same domain samples are divided into easier domains such as \textit{Photo}. The \textit{Sketch} distribution is affected by a more consistent domain gap, resulting in a more agglomerate domain cluster of samples. From this representation, the competitive advantage of transformer-based backbones is especially evident for \textit{Art Painting}, although valuable in the classification accuracy on every target domain. The model marked with * is pretrained on Imagenet21K instead of ImageNet1K.}
  \label{fig:tSNE_full}
\end{figure}

\cref{sec:model_introspections} reports and discusses the visualization of the PACS domain \textit{Art Painting} with ResNet50 and ConViT, highlighting the advantage of using transformer-based networks. In this section, we propose an additional t-SNE single-domain representation of features extracted from all PACS domains, with ResNet50 and our three best backbones (\cref{fig:tSNE_full}). According to the higher distance between source and target distributions, more challenging target domains result in more agglomerate clusters of domain samples. From this representation, the competitive advantage offered by transformer-based backbones is especially evident for \textit{Art Painting}. ConViT shows more separated class features for the \textit{Cartoon} domain.  These findings confirm the baseline results reported in \cref{sec:baseline_benchmark}, in which transformers show valuable improvements on every target domain.

\subsection{Self-attention Visualization}
\begin{figure}[ht]
    \centering
    \includegraphics[width=\linewidth]{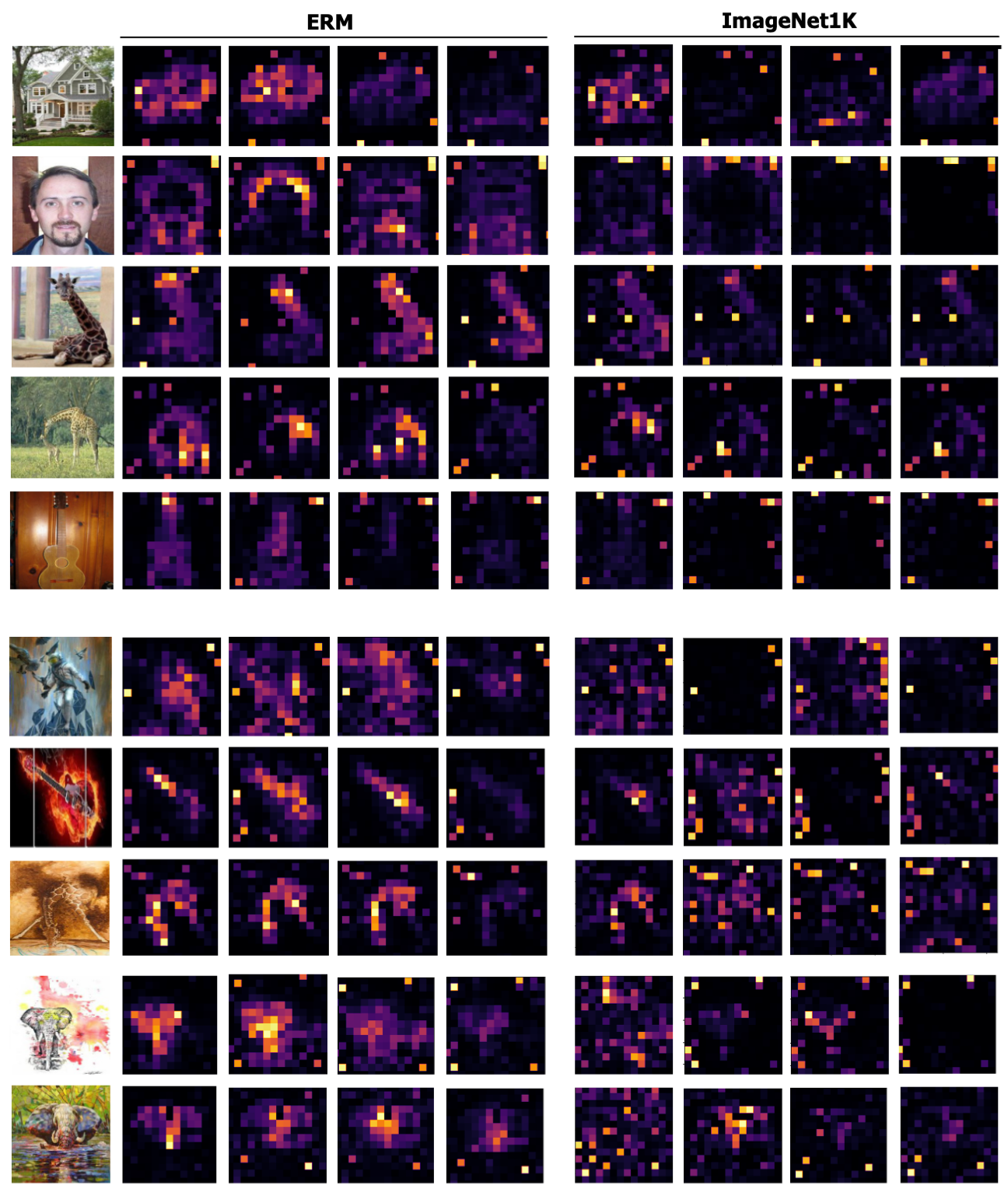}
    \caption{Self-attention DeiT Base of most active heads of the last layer for some samples of the \textit{Photo} and \textit{Art Painting} PACS domains. We look at the attention map when using the {\fontfamily{qcr}\selectfont [CLS]} token as a query for the different heads in the last layer. It is clear how ERM is very effective at effectively redirecting attention toward more meaningful regions and mitigating pretraining noise.}
    \label{fig:additional_self_attention_1}
\end{figure}

\begin{figure}[ht]
    \centering
    \includegraphics[width=\linewidth]{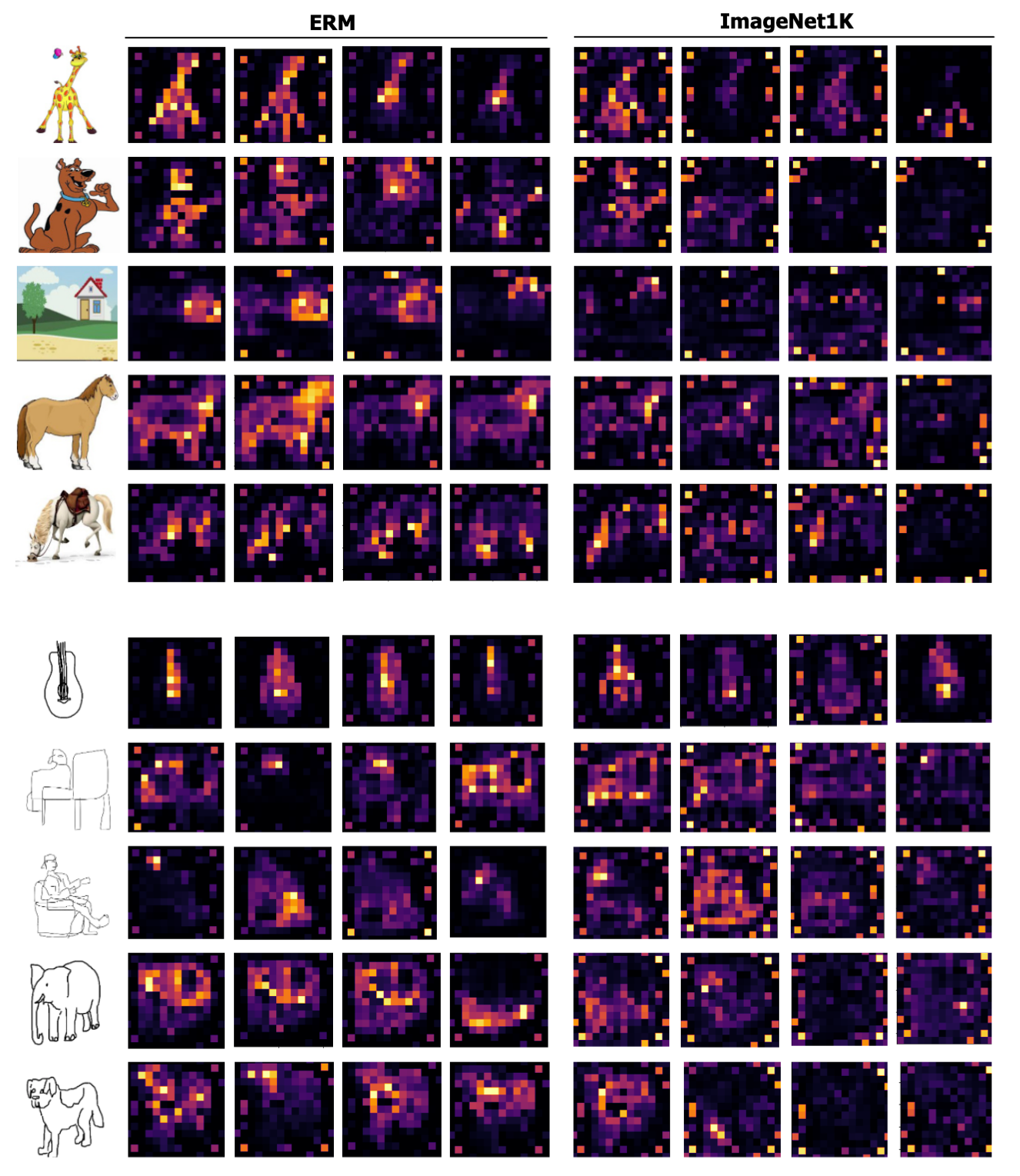}
    \caption{Self-attention DeiT Base of most active heads of the last layer for some samples of the \textit{Cartoon} and \textit{Sketch} PACS domains. We look at the attention map when using the {\fontfamily{qcr}\selectfont [CLS]} token as a query for the different heads in the last layer. It is clear how ERM is very effective at effectively redirecting attention toward more meaningful regions and mitigating pretraining noise.}
    \label{fig:additional_self_attention_2}
\end{figure}

We provide more self-attention visualizations for randomly selected PACS images: \textit{Photo} and \textit{Art Painting} domains in \cref{fig:additional_self_attention_1}, \textit{Cartoon} and \textit{Sketch} in \cref{fig:additional_self_attention_2}. We show the four most active heads of DeiT Base using the {\fontfamily{qcr}\selectfont [CLS]} token as a query for the different heads of the last layer. It is clear how ERM maps present more localized attention regions, focusing on more meaningful features. Finally, highly active isolated patches are learned during ImageNet training due to overfitting; even if some pretraining noise remains, ERM strongly attenuates this problem, further focalizing the attention of the network and reducing biased predictions.

\clearpage
\end{document}